\pdfoutput=1
\documentclass[11pt]{article}

\usepackage[]{acl}

\usepackage{times}
\usepackage{latexsym}

\usepackage[T1]{fontenc}
\usepackage{booktabs}

\usepackage{multirow}
\usepackage{adjustbox}
\usepackage{amsmath}

\usepackage{subcaption}
\usepackage{makecell}
\usepackage{color}
\usepackage{tikz}
\usepackage{pgf}
\usepackage{pgfplots}
\usetikzlibrary{patterns}
\usepackage{tikz-qtree}
\usetikzlibrary{arrows,decorations.pathmorphing,backgrounds,positioning,fit,petri,shapes.misc, arrows.meta,shapes.geometric,decorations.markings,calc,shadows.blur,decorations.pathreplacing,quotes,matrix,shapes.symbols}
\usepackage{amssymb}

\usepackage{fdsymbol}

\usetikzlibrary{arrows,decorations.pathmorphing,backgrounds,positioning,fit,petri,shapes.misc, arrows.meta,shapes.geometric,decorations.markings,calc,shadows.blur,decorations.pathreplacing,quotes,matrix,shapes.symbols}
\tikzset{
table/.style={
first column text width/.code={%
	\tikzset{
		column 1/.style={
			nodes={text width=##1, font=\bfseries}
		},
	}
},
first column text width=5em
}
}

\definecolor{fontgray}{RGB}{44, 62, 80}
\definecolor{myred}{RGB}{235, 47, 6} %rgb()
\definecolor{summertime}{RGB}{245, 205, 121}
\definecolor{darkgrass}{RGB}{0, 148, 50}
\definecolor{myblue}{RGB}{0, 168, 255}
\definecolor{mygray}{RGB}{158, 158, 158}
\definecolor{puffin}{RGB}{250, 152, 58}
\definecolor{lowpurple}{RGB}{210, 180, 222}
\definecolor{lowblue}{RGB}{102,178,255}
\definecolor{lowred}{RGB}{245, 183, 177}
\definecolor{deeppurple}{RGB}{142, 68, 173}
\definecolor{nephritis}{RGB}{39, 174, 96}

\definecolor{deepblue}{RGB}{41, 128, 185}
\definecolor{shymoment}{RGB}{162, 155, 254}
\definecolor{firstdate}{RGB}{250, 177, 160}
\definecolor{mintleaf}{RGB}{0, 184, 148}
\definecolor{alizarin}{RGB}{231, 76, 60}
\definecolor{soaring}{RGB}{149, 175, 192}
\definecolor{electronblue}{RGB}{9, 132, 227}
\definecolor{pinkgla}{RGB}{0, 184, 148}

\renewcommand{\vec}[1]{\mathbf{#1}}

\DeclareMathOperator*{\ffn}{FFN}
\DeclareMathOperator*{\rationalizer}{Rationalizer}

\tikzset{nomorepostaction/.code=\let\tikz@postactions\pgfutil@empty}
\usepackage{pifont}% http://ctan.org/pkg/pifont
\newcommand{\cmark}{\ding{51}}%
\newcommand{\xmark}{\ding{55}}%
\def\Section {\S}
\usepackage[utf8]{inputenc}

\usepackage{microtype}

\newcommand{\squishlist}{
\begin{list}{$\bullet$}
{ \setlength{\itemsep}{0pt}
	\setlength{\parsep}{3pt}
	\setlength{\topsep}{3pt}
	\setlength{\partopsep}{0pt}
	\setlength{\leftmargin}{1.5em}
	\setlength{\labelwidth}{1em}
	\setlength{\labelsep}{0.5em} } }

\newcounter{Lcount}
\newcommand{\squishlisttwo}{
\begin{list}{\arabic{Lcount}. }
	{ \usecounter{Lcount}
		\setlength{\itemsep}{0pt}
		\setlength{\parsep}{0pt}
		\setlength{\topsep}{0pt}
		\setlength{\partopsep}{0pt}
		\setlength{\leftmargin}{2em}
		\setlength{\labelwidth}{1.5em}
		\setlength{\labelsep}{0.5em} } }

\newcommand{\squishend}{\end{list} }

\title{Learning to Reason Deductively: \\ Math Word Problem Solving as Complex Relation Extraction}

\author{Zhanming Jie$^{\varheartsuit\vardiamondsuit}$, Jierui Li$^{\clubsuit\vardiamondsuit}$ \and Wei Lu$^{\vardiamondsuit}$ \\
$^{\varheartsuit}$ByteDance AI Lab,  ~$^{\clubsuit}$University of Texas at Austin \\
$^{\vardiamondsuit}$StatNLP Research Group, Singapore University of Technology and Design\\
\texttt{allan@bytedance.com}, \texttt{jierui@cs.utexas.edu}, \texttt{luwei@sutd.edu.sg} \\
}

\begin{document}
\maketitle
\begin{abstract}
	Solving math word problems requires deductive reasoning over the quantities in the text.
	Various recent research efforts mostly relied on sequence-to-sequence or sequence-to-tree models to generate mathematical expressions without explicitly performing relational reasoning between quantities in the given context.
	While empirically effective, such approaches typically do not provide explanations for the generated expressions.
	In this work, we view the task as a {\em complex relation extraction} problem,  proposing a novel approach that presents explainable deductive reasoning steps to iteratively construct target expressions, where each step involves a primitive operation over two quantities defining their relation.
	Through extensive experiments on four benchmark datasets, we show that the proposed model significantly outperforms existing strong baselines. 
	We further demonstrate that the deductive procedure not only presents more explainable steps but also enables us to make more accurate predictions on questions that require more complex reasoning.\footnote{Our code and data are released at \url{https://github.com/allanj/Deductive-MWP}.}
\end{abstract}
%% presented in the text
%% remove "carefully", stesp which are more explainable, two different languea
%%  across two languages, is able to significantly ot perform

\section{Introduction}
\label{sec:intro}
Math word problem (MWP) solving~\cite{bobrow1964natural} is a task of answering a mathematical question that is described in natural language.
Solving MWP requires logical reasoning over the quantities presented in the context~\cite{mukherjee2008review} to compute the numerical answer.
Various recent research efforts regarded the problem as a generation problem -- typically, such models focus on generating the complete target mathematical expression, often represented in the form of a linear sequence or a tree structure~\cite{xie2019goal}.

\begin{figure}[t!]
	\centering
	\adjustbox{max width=0.95\linewidth}{
		\begin{tikzpicture}[node distance=2.0mm and 2.0mm, >=Stealth, 
			wordnode/.style={draw=none, minimum height=5mm, inner sep=0pt},
			chainLine/.style={line width=0.8pt,-, color=mygray},
			entbox/.style={draw=black, rounded corners, fill=red!20, dashed},
			mathop/.style={draw=none, circle, minimum height=2mm, inner sep=1pt, line width=0.8pt, fill=pinkgla},
			stepsty/.style={draw=pinkgla, circle, minimum height=2mm, inner sep=1pt, line width=0.8pt},
			quant/.style={draw=none, minimum height=2mm, inner sep=1pt, line width=1pt},
			expr/.style={draw=electronblue, rectangle, minimum width=20mm, minimum height=4.5mm, inner sep=2pt, line width=0.8pt, rounded corners},
			box/.style={draw=electronblue, rectangle, rounded corners, line width=0.9pt, dashed, minimum width=26mm, minimum height=20mm},
			invis/.style={draw=none},
			]
			
			\node[wordnode, align=left] (question) [] {\textbf{Question}: \it In a division sum , the remainder is \textbf{8}\\\it and the divisor is \textbf{6} times the quotient and is obt-\\\it-ained by adding \textbf{3} to the thrice of the remainder.\\\it What is the dividend?};
			\node[wordnode, below=of question.west,anchor=west, yshift=-11mm] (answer) [] {\small \textbf{Answer}:  $129.5$} ;
			\node[wordnode, right=of answer, xshift=-1mm] (expression) [] {\small \textbf{Expr}:  $\big ( \mathbf{\textcolor{electronblue}{(8\times 3 + 3)}}\! \times \!  \mathbf{\textcolor{electronblue}{(8 \times 3 + 3)}}   \!\div\! 6\big )\! +\! 8$} ;

			\node[wordnode] (model1) [below= of answer, xshift=7mm] {Tree generation: $7$ ops};
			%		\node[wordnode] (model11) [below= of model1, yshift=7mm] {\Large Generation};
			\node[mathop, fill=pinkgla, draw=none] (t00) [below= of expression, xshift=5mm, yshift=1mm] {$\mathbf{\textcolor{white}{+}}$};
			\node[mathop] (t10) [below left= of t00, xshift=-7mm] {$\mathbf{\textcolor{white}{\times}}$};
			\node[quant] (t11) [below right= of t00, xshift=3mm] {${\textcolor{black}{8}}$};
			\node[mathop] (t20) [below left= of t10, xshift=-10mm] {$\mathbf{\textcolor{white}{+}}$};
			\node[mathop] (t21) [below right= of t10, xshift=8mm] {$\mathbf{\textcolor{white}{\div}}$};
			\node[mathop] (t30) [below left= of t20, xshift=-4mm, yshift=-3mm] {$\mathbf{\textcolor{white}{\times}}$};
			\node[quant] (t31) [below right= of t20, xshift=3mm] {${\textcolor{black}{3}}$};
			\node[mathop] (t32) [below left=  of t21] {$\mathbf{\textcolor{white}{+}}$};
			\node[quant] (t33) [below right= of t21, xshift=4mm] {${\textcolor{black}{6}}$};
			\node[quant] (t40) [below left=  of t30] {${\textcolor{black}{8}}$};
			\node[quant] (t41) [below right= of t30] {${\textcolor{black}{3}}$};
			\node[mathop] (t42) [below left=  of t31, xshift=15mm] {$\mathbf{\textcolor{white}{\times}}$};
			\node[quant] (t43) [below right= of t31, xshift=17mm] {${\textcolor{black}{3}}$};
			\node[quant] (t50) [below left=  of t42] {${\textcolor{black}{8}}$};
			\node[quant] (t51) [below right= of t42] {${\textcolor{black}{3}}$};
			
			\node[invis] (bodera)[below= of answer, yshift=-30mm, xshift=-8mm] {};
			\node[invis] (boderb)[right= of bodera, xshift=65mm] {};
			\draw [chainLine,  dashed, fontgray] (bodera) to [] node[] {} (boderb);
			%		
			%		\node[wordnode] (opnum1) [below= of t50, yshift=2mm] {\Large \#operations: 7};
			%				
			\draw [chainLine, ->] (t00) to [out=-140,in=20] node[] {} (t10);
			\draw [chainLine, ->] (t00) to [out=-60,in=170] node[] {} (t11);
			
			\draw [chainLine, ->] (t10) to [out=-140,in=20] node[] {} (t20);
			\draw [chainLine, ->] (t10) to [out=-60,in=170] node[] {} (t21);
			
			\draw [chainLine, ->] (t20) to [out=-140,in=30] node[] {} (t30);
			\draw [chainLine, ->] (t20) to [out=-60,in=170] node[] {} (t31);
			\draw [chainLine, ->] (t21) to [] node[] {} (t32);
			\draw [chainLine, ->] (t21) to [out=-60,in=170] node[] {} (t33);
			
			\draw [chainLine, ->] (t30) to [] node[] {} (t40);
			\draw [chainLine, ->] (t30) to [] node[] {} (t41);
			\draw [chainLine, ->] (t32) to [] node[] {} (t42);
			\draw [chainLine, ->] (t32) to [out=-60,in=160] node[] {} (t43);
			\draw [chainLine, ->] (t42) to [] node[] {} (t50);
			\draw [chainLine, ->] (t42) to [] node[] {} (t51);
			\begin{pgfonlayer}{background}
				\node [box] (dup1) [below=of model1, xshift=5mm,yshift=-2mm] {};
				\node [box, minimum width=21mm, minimum height=16mm, yshift=2mm] (dup2) [right=of dup1, yshift=-5mm, xshift=-1mm] {};
			\end{pgfonlayer}

			%% Expression Parsing approach.
			
			\node[wordnode] (model21) [below = of t40, xshift=13mm,yshift=-2mm] {Our deductive procedure: $5$ ops};
			
			%		\node[quant] (m01) [right= of m00, xshift=10mm] {\Large\textbf{\textcolor{deeppurple}{6}}};
			%		\node[quant] (m02) [right= of m01, xshift=10mm] {\Large\textbf{\textcolor{deeppurple}{3}}};
			%		\node[quant] (m03) [right= of m02, xshift=10mm] {\Large\textbf{\textcolor{deeppurple}{3}}};
			%		
			\node[expr] (m10) [below= of t40, yshift=-7mm, xshift=2mm] {\small \textcolor{black}{$8 \times 3 = 24$}};
			\node[stepsty] (st1) [below= of m10, yshift=1mm] {\small  \textcolor{pinkgla}{$1$}};
			\node[expr] (m20) [right= of m10, xshift=4mm] {\small  {\textcolor{black}{$24 + 3 = 27$}}};
			\node[stepsty] (st2) [below= of m20, yshift=1mm] {\small  \textcolor{pinkgla}{$2$}};
			\node[expr] (m30) [right= of m20, xshift=4mm] {{\textcolor{black}{\small  $27 \div 6 = 4.5$}}};
			\node[stepsty] (st3) [below= of m30, yshift=1mm] {\small  \textcolor{pinkgla}{$3$}};
			
			\node[expr] (m40) [below left= of m20, xshift=10mm, yshift=-5mm] {\small  {\textcolor{black}{$27 \times 4.5 = 121.5$}}};
			\node[stepsty] (st4) [below= of m40, yshift=1mm] {\small  \textcolor{pinkgla}{$4$}};
			\node[expr] (m50) [right= of m40, xshift=4mm, yshift=0mm] {\small  {\textcolor{black}{$121.5 + 8 = 129.5$}}};
			\node[stepsty] (st5) [below= of m50, yshift=1mm] {\small  \textcolor{pinkgla}{$5$}};
			%		\node[wordnode] (opnum2) [below= of m01, yshift=2mm] {\Large \#operations: 5};
			%		
			%		\node[wordnode] (model2) [right = of m50] {\Large Expression};
			%		\node[wordnode] (model21) [below = of model2, yshift=6mm] {\Large Parsing};
			%		
			%		
			%		\draw [chainLine, ->] (m00) to [] node[] {} (m10);
			%		\draw [chainLine, ->] (m02) to [] node[] {} (m10);
			%		
			\draw [chainLine, ->] (m10) to [] node[] {} (m20);
			%		\draw [chainLine, ->] (m03) to [] node[] {} (m20);
			%		
			%		\draw [chainLine, ->] (m01) to [] node[] {} (m30);
			\draw [chainLine, ->] (m20) to [] node[] {} (m30);
			\draw [chainLine, ->] (m30) to [] node[] {} (m40);
			\draw [chainLine, ->] (m20) to [] node[] {} (m40);
			\draw [chainLine, ->] (m40) to [] node[] {} (m50);
			%		
			%		\draw [chainLine, ->] (m00) to [out=120,in=-165, looseness=1] node[] {} (m50);
			
		\end{tikzpicture}
	}
	\vspace{-2mm}
	\caption{A MWP example taken from  MathQA. 
		Top: tree generation.  
		Bottom:  deductive procedure.}
	\label{fig:example}
\end{figure}

Figure \ref{fig:example} (top) depicts a typical approach that attempts to generate the target expression in the form of a tree structure, which is adopted in recent research efforts~\cite{xie2019goal,zhang2020graph,patel2021are,wu2021math}.
Specifically, the output is an expression that can be obtained from such a generated structure.
We note that, however, there are several limitations with such a structure generation approach.
First, such a process typically involves a particular order when generating the structure.
In the example, given the complexity of the problem, the decision of generating the addition (``$+$'') operation as the very first step could be counter-intuitive and does not provide adequate explanations that show the reasoning process when being presented to a human learner.
Furthermore, the resulting tree contains identical sub-trees (``$8\times3+3$'') as highlighted in blue dashed boxes.
Unless a certain specifically designed mechanism is introduced for reusing the already generated intermediate expression, the approach would need to repeat the same effort in its process for generating the same sub-expression.

Solving math problems generally requires deductive reasoning, which is also regarded as one of the important abilities in children's cognitive development \cite{piaget1952child}.
In this work, we propose a novel approach that explicitly presents deductive reasoning steps.
We make a key observation that MWP solving fundamentally can be viewed as a {\em complex relation extraction} problem -- the task of identifying the complex relations
among the quantities that appear in the given problem text.
Each primitive arithmetic operation (such as {\em addition}, {\em subtraction}) essentially defines a different type of relation.
Drawing on the success of some recent models for relation extraction in the literature \cite{zhong-chen-2021-frustratingly},
our proposed approach involves a process that repeatedly performs relation extraction between two chosen quantities (including newly generated quantities).

As shown in Figure \ref{fig:example}, our approach directly extracts the relation (``{\em multiplication}'', or ``$\times$'') between $8$ and $3$, which come from the contexts ``\textit{remainder is 8}'' and ``\textit{thrice of the remainder}''.
In addition, it allows us to reuse the results from the intermediate expression in the fourth step.
This process naturally yields a deductive reasoning procedure that iteratively derives new knowledge from existing ones.
Designing such a complex relation extraction system presents several practical challenges.
% For example, some quantities that appear in the text may be irrelevant to the question while some others may need to be used multiple times.
For example, some quantities may be irrelevant to the question while some others may need to be used multiple times.
The model also needs to learn how to properly handle the new quantities that emerge from the intermediate expressions.
Learning how to effectively search for the optimal sequence of operations (relations) and when to stop the deductive process is also important.

In this work, we tackle the above challenges and make the following major contributions:
\squishlist
\item We formulate MWP solving as a complex relation extraction task,
where we aim to repeatedly identify the basic relations between different quantities.
To the best of our knowledge, this is the first effort that successfully tackles MWP solving from such a new perspective.
\item Our model is able to automatically produce explainable steps that lead to the final answer, presenting a deductive reasoning process.
\item Our experimental results on four standard datasets across two languages show that our model significantly outperforms existing strong baselines.
We further show that the model performs better on problems with more complex equations than previous approaches.
\squishend

\section{Related Work}

% \paragraph{Math Word Problem Solving}
Early efforts focused on solving MWP using probabilistic models with handcrafted features~\cite{liguda2012modeling}. 
\citet{kushman2014learning} and \citet{roy2018mapping} designed templates to find the alignments between the declarative language and equations.
Most recent works solve the problem by using sequence or tree generation models.
\citet{wang2017deep} proposed the Math23k dataset and presented a sequence-to-sequence (seq2seq) approach to generate the mathematical expression~\cite{chiang-chen-2019-semantically}. 
Other approaches improve the seq2seq model with reinforcement learning~\cite{huang2018neural}, template-based methods~\cite{wang2019template}, and group attention mechanism~\cite{li2019modeling}.
\citet{xie2019goal} proposed a goal-driven tree-structured (GTS) model to generate the expression tree. 
This sequence-to-tree approach significantly improved the performance over the traditional seq2seq approaches. 
Some follow-up works incorporated external knowledge such as syntactic dependency~\cite{shen2020solving,lin2021hms} or commonsense knowledge~\cite{wu2020knowledge}.
\citet{cao2021bottom} modeled the equations as a directed acyclic graph to obtain the expression.
\citet{zhang2020graph} and \citet{li2020graph} adopted a graph-to-tree approach to model the quantity relations using the graph convolutional networks (GCN)~\cite{kipf2017semi}.
Applying pre-trained language models such as BERT~\cite{devlin2019bert} was shown to significantly benefit the tree expression generation~\cite{lan2021mwptoolkit,tan2021investigating,liang2021mwp,li2021seeking,shen2021generate}. 

Different from the tree-based generation models, our work is related to deductive systems~\cite{shieber1995principles,nederhof2003weighted} where we aim to obtain step-by-step expressions. 
Recent efforts have also been working towards this direction.
\citet{ling2017program} constructed a dataset to provide explanations for expressions at each step. 
\citet{amini2019mathqa} created the MathQA dataset annotated with step-by-step operations.
The annotations present the expression at each intermediate step during problem-solving. 
Our deductive process (Figure \ref{fig:example}) attempts to automatically obtain the expression in an incremental, step-by-step manner.

% \paragraph{Relation Extraction}
Our approach is also related to relation extraction (RE) ~\cite{zelenko2003kernel}, a fundamental task in the field of information extraction that is focused on identifying the relationships between a pair of entities.
Recently, \citet{zhong-chen-2021-frustratingly} designed a simple and effective approach to directly model the relations on the span pair representations. 
In this work, we treat the operation between a pair of quantities as the relation at each step in our deductive reasoning process. 
Traditional methods~\cite{liang2018meaning} applied rule-based approaches to extract the mathematical relations.

% \paragraph{Reasoning}
MWP solving is typically regarded as one of the {\em system 2} tasks \cite{kahneman2011thinking,bengio2021deep}, and our current approach to this problem is related to neural symbolic reasoning \cite{besold2017neural}.
We design differentiable modules~\cite{andreas2016neural,gupta2020neural} in our model (\Section\ref{sec:parsing}) to perform reasoning among the quantities.
% {\color{blue}Specifically, the modules represent the mathematical operators that operate on the quantities to produce expressions.}

\begin{figure}[t!]
	\centering
	\adjustbox{max width=0.95\linewidth}{
		\begin{tikzpicture}[node distance=2.0mm and 2.0mm, >=Stealth, 
			wordnode/.style={draw=none, minimum height=5mm, inner sep=0pt},
			chainLine/.style={line width=0.8pt,-, color=mygray},
			entbox/.style={draw=black, rounded corners, fill=red!20, dashed},
			mathop/.style={draw=deepblue, circle, minimum height=2mm, inner sep=1pt, line width=0.8pt},
			quant/.style={draw=none, minimum height=2mm, inner sep=1pt, line width=1pt},
			expr/.style={draw=deepblue, rectangle, minimum width=20mm, inner sep=2pt, line width=0.8pt, rounded corners},
			box/.style={draw=alizarin, rectangle, rounded corners, line width=0.9pt, dashed, minimum width=28mm, minimum height=22mm},
			invis/.style={draw=none},
			]
			
			%		%% In a division sum , the remainder is 8 and the divisor is 6 times 
			%     the quotient and is obtained by adding 3 to the thrice of the remainder. What is the divident?
			\matrix (sent) [table, matrix of nodes, column 1/.style={anchor=base east}, column 2/.style={anchor=base west}, nodes in empty cells, first column text width=3em, execute at empty cell=\node{\strut};]
			{
				input: & $q$ in $\mathcal{Q}^{(0)}$ \\ 
				axiom: & $0: \langle q_1, \cdots, q_{|\mathcal{Q}^{(0)}|}  \rangle$\\
				& $t: \big \langle q_1, \cdots, q_{|\mathcal{Q}^{(t-1)}|} \big \rangle$ \\
				&  $t+1: \big \langle q_1, \cdots, q_{|\mathcal{Q}^{(t-1)}|} ~\vert ~  q_{|\mathcal{Q}^{(t)}|}:= e_{i,j,op}^{(t)} \big \rangle$ \\
			};
			\node[fit=(sent-3-1)(sent-4-1), xshift=-2mm] (op){$q_i \xrightarrow[]{op} q_j$:};
			\node[invis, right=of op, yshift=1mm, xshift=-5mm] (a) {};
			\node[invis, right= of a, xshift=63mm] (b){};
			\draw[chainLine, -, line width=1pt] (a) to (b);
		\end{tikzpicture}
	}
	\vspace{-3mm}
	\caption{Our deductive system. $t$ is the current step. $\langle\cdot\rangle$ denotes the quantity list.}
	\label{fig:deductive_system}
\end{figure}

\section{Approach}

The math word problem solving task can be defined as follows.
{\color{black}Given a problem description $ \mathcal{S} = \{w_1, w_2, \cdots, w_n\}$ that consists of a list of $n$ words and $\mathcal{Q}_S = \{q_1, q_2, \cdots, q_m\}$, a list of $m$ quantities that appear in $\mathcal{S}$, 
	our task is to solve the problem and return the numerical answer.}
Ideally, the answer shall be computed through a mathematical reasoning process over a series of primitive mathematical operations~\cite{amini2019mathqa} as shown in Figure \ref{fig:example}.
Such operations may include ``$+$'' ({\em addition}), ``$-$'' ({\em subtraction}), ``$\times$'' ({\em multiplication}), ``$\div$'' ({\em division}), and ``$**$'' ({\em exponentiation}).\footnote{While we consider binary operators, extending our approach to support unary or ternary operators is possible (\Section \ref{sec:discussion}).}

In our view, each of the primitive mathematical operations above can essentially be used for describing a specific {\em relation} between quantities.
Fundamentally, solving a math word problem is a problem of {\em complex relation extraction}, which requires us to repeatedly identify the relations between quantities (including those appearing in the text and those intermediate ones created by relations).
The overall solving procedure requires invoking a relation classification module at each step, yielding a deductive reasoning process.

In practice, some questions cannot be answered without relying on certain predefined constants (such as $\pi$ and $1$) that may not have appeared in the given problem description.
We therefore also consider a set of constants $\mathcal{C} = \{c_1, c_2, \cdots, c_{\vert \mathcal{C}\vert }\}$.
Such constants are also regarded as quantities (i.e., they would be regarded as $\{q_{m+1},q_{m+2},\dots,q_{m+|\mathcal{C}|}\}$) which may play useful roles when forming the final answer expression.

\subsection{A Deductive System}

As shown in Figure \ref{fig:example}, applying the mathematical relation (e.g., ``$+$'') between two quantities yields an intermediate expression $e$. 
In general, at step $t$, the resulting expression $e^{(t)}$ (after evaluation) becomes a {\em newly created quantity} that is added to the list of candidate quantities and is ready for participating in the remaining deductive reasoning process from step $t+1$ onward.
This process can be mathematically denoted as follows:
%\begin{itemize}
\squishlist
\item Initialization:
% \begin{equation}
	%     \mathcal{Q}^{(0)} = \mathcal{Q}_S \cup \mathcal{C}  \nonumber
	% \end{equation}
% \begin{center}
	% 		\begin{tabular}{c}
		% 			$
		% 			\mathcal{Q}^{(0)} = \mathcal{Q}_S \cup \mathcal{C}  
		% 			$
		% 		\end{tabular}
	% 		%\vspace{-2mm}
	% 	\end{center}
\begin{eqnarray}
	\mathcal{Q}^{(0)} &=& \mathcal{Q}_{\mathcal{S}} \cup \mathcal{C} ~~~~~~~~~~~~~~~~~~~~~~~~~~\nonumber
\end{eqnarray}
\item \text{At step $t$:}
% \begin{equation}
	%     \begin{split}
		%         e_{i,j,op}^{(t)} & = & q_i \xrightarrow[]{op} q_j ~~~q_i, q_j  \in \mathcal{Q}^{(t-1)}\\
		%         \mathcal{Q}^{(t)} & = & \mathcal{Q}^{(t-1)} \cup \{e_{i,j,op}^{(t)}\} \\
		%         q_{|\mathcal{Q}^{(t)}|} & := & e_{i,j,op}^{(t)} \\
		%     \end{split}\nonumber
	% \end{equation}
\begin{eqnarray}
	e_{i,j,op}^{(t)} &=& q_i \xrightarrow[]{op} q_j ~~~q_i, q_j  \in \mathcal{Q}^{(t-1)}\nonumber
	\\
	\mathcal{Q}^{(t)} &=& \mathcal{Q}^{(t-1)} \cup \{e_{i,j,op}^{(t)}\} \nonumber
	\\
	q_{|\mathcal{Q}^{(t)}|} &:=& e_{i,j,op}^{(t)}\nonumber
\end{eqnarray}
%\end{itemize}
\squishend
where $e_{i,j,op}^{(t)}$ represents the expression after applying the relation $op$ to the ordered pair $(q_i, q_j)$.
Following the standard deduction systems~\cite{shieber1995principles,nederhof2003weighted}, the reasoning process can be formulated in Figure \ref{fig:deductive_system}.
We start with an axiom with the list of quantities in $\mathcal{Q}^{(0)}$. 
The inference rule is $q_i \xrightarrow[]{op} q_j$ as described above to obtain the expression as a new quantity at step $t$. 
%After each time step, the state accepts the resulting expression as a new quantity for following calculation. {\color{red}not clear at all. Need more explanations to the Figure and what it means..}

\definecolor{coral}{RGB}{255, 127, 80}

\subsection{Model Components}
\label{sec:parsing}

\begin{figure}[t!]
	\centering
	\adjustbox{max width=1\linewidth}{
		\begin{tikzpicture}[node distance=1.0mm and 5.0mm, >=Stealth, 
			wordnode/.style={draw=none, minimum height=5mm, inner sep=0pt},
			chainLine/.style={line width=1.1pt,-, color=deepblue},
			entbox/.style={draw=black, rounded corners, fill=red!20, dashed},
			mathop/.style={draw=deepblue, circle, minimum height=7mm, inner sep=1pt, line width=1pt},
			quant/.style={draw=deeppurple, circle, minimum height=7mm, inner sep=1pt, line width=1pt},
			expr/.style={draw=none, rectangle, minimum width=10mm, inner sep=1pt, line width=1pt, rounded corners, minimum height=4mm, fill=puffin!40},
			invis/.style={draw=none},
			plm/.style={draw=none, fill=lowblue!60, rectangle, rounded corners, minimum width=140mm, minimum height=9mm},
			score/.style={draw=none, fill=myred!40, circle, minimum width=5mm},
			vector/.style={draw=none, fill=coral!60, minimum height=4mm, minimum width=10mm, inner sep=1pt}
			]
			
			%			\node[wordnode] (example) {\textit{If a machine can make 2,088 gears in 8 hours, how many gears it make in 9 hours?}};
			%			\node[plm]  (bert) [] {\textcolor{fontgray}{Pre-trained Language Model }};
			%			%				\node[wordnode, below=of bert] (sent) [] {A machine can make \textbf{2088} gears in \textbf{}8 hours, How many gears can it make in 9 hours?};
			%			%				If a machine can make 2088 gears in 8 hours , How many gears can it make in 9 hours ?
			\matrix (sent) [matrix of nodes, nodes in empty cells, execute at empty cell=\node{\strut};, row sep=3mm]
			{
				\textit{If} &[-1mm] \textit{a} &[-1mm] \textit{machine} &[-1mm] \textit{can} & [-1mm] \textit{make}   &    [-1mm] \textit{\textbf{\underline{2,088}}} &[-1mm] \textit{gears} & [-1mm] \textit{in} & [-1mm] \textit{\textbf{\underline{8}}} & [-1mm] \textit{hours,} \\
				\textit{how}  & [-1mm]
				\textit{many} &[-1mm] \textit{gears} &[-1mm] \textit{it} &[-1mm] \textit{make} &[-1mm] \textit{in} &[-1mm] \textit{\textbf{\underline{9}}} &[-1mm] \textit{hours?} \\
				%		 \textbf{\textsc{date}}   &   &   & \textsc{o}     & \textsc{o}   & \textsc{o}  & \textsc{o}& \textsc{event} &  \\
			};
			\node[wordnode, below=of sent-1-6, yshift=2mm] (v1) [] {\textcolor{myred}{$q_1$}};
			\node[wordnode, below=of sent-1-9, yshift=2mm] (v2) [] {\textcolor{myred}{$q_2$}};
			\node[wordnode, below=of sent-2-7, yshift=2mm] (v3) [] {\textcolor{myred}{$q_3$}};

			\node[vector, below=of sent-2-2, yshift=-5mm] (v1v)[]{$\boldsymbol{q}_1$}; 
			\node[vector, right=of v1v, xshift=10mm] (v2v)[]{$\boldsymbol{q}_2$}; 
			\node[vector, right=of v2v, xshift=10mm] (v3v)[]{$\boldsymbol{q}_3$}; 
			
			\node[wordnode, below left= of v1v, xshift=3mm] (t0) []{$t=1$};
			
			\node[expr, below right=of v1v, fill=shymoment!40, xshift=-12mm, yshift=-9mm, minimum width=15mm, minimum height=6mm] (v12) [] {$\left [ \boldsymbol{q}_1, \boldsymbol{q}_2, \boldsymbol{q}_1 \circ \boldsymbol{q}_2 \right ]$};
			
			\draw [chainLine, ->, mygray, line width=0.8pt] (v1v) to (v12);
			\draw [chainLine, ->, mygray, line width=0.8pt] (v2v) to (v12);
			
			\node[right=of v12, fill=darkgrass!40, minimum width=15mm, minimum height=6mm,yshift=-3mm] (op) [] {FFN$_{op=\text{``}\div\text{''}}$};
			
			\node[right=of v12, fill=darkgrass!20, minimum width=15mm, minimum height=6mm,yshift=5mm] (opwrong) [] {FFN$_{op=\text{``}\times\text{''}}$};
			
			\draw [chainLine, ->, mygray, line width=0.8pt] (v12) to (op);
			\draw [chainLine, ->, mygray!50, line width=0.8pt] (v12) to (opwrong);
			
			\node[expr, right =of op, minimum width=15mm, minimum height=6mm, xshift=5mm, label=right:{\Large \textcolor{nephritis}{\cmark}}] (m1) [] {$\boldsymbol{e}_{1,2,\div}$};
			\node[expr, right =of opwrong, minimum width=15mm, minimum height=6mm, xshift=5mm, fill=puffin!10, label=right:{\Large \textcolor{myred}{\xmark}}] (m1wrong) [] {$\boldsymbol{e}_{1,2,\times}$};
			%			, label=above:{$e_{1,2,/}^{(1)}$}
			\draw [chainLine, ->, mygray, line width=0.8pt] (op) to (m1);
			
			\draw [chainLine, ->, mygray!30, line width=0.8pt] (opwrong) to (m1wrong);

			\node[invis] (bodera)[below= of v1v, yshift=-19mm, xshift=-10mm] {};
			\node[invis] (boderb)[right= of bodera, xshift=74mm] {};
			\draw [chainLine,  dash dot, fontgray] (bodera) to [] node[] {} (boderb);

			\node[vector, below=of v1v, yshift=-25mm] (v1vv)[]{$\boldsymbol{q}_1^{\prime}$}; 
			\node[vector, right=of v1vv, xshift=6mm] (v2vv)[]{$\boldsymbol{q}_2^{\prime}$}; 
			\node[vector, right=of v2vv, xshift=6mm] (v3vv)[]{$\boldsymbol{q}_3^{\prime}$};
			\node[vector, right=of v3vv, xshift=6mm] (r1)[]{$\boldsymbol{q}_4$}; 
			
			\node[wordnode, below left= of v1vv, xshift=3mm] (t1) []{$t=2$};
			
			\node[expr, below right=of v1vv, fill=shymoment!40, xshift=-12mm, yshift=-6mm, minimum width=15mm, minimum height=6mm] (m1v3) [] { $\left [ \boldsymbol{q}_{3}^{\prime}, \boldsymbol{q}_4, \boldsymbol{q}_{3}^{\prime} \circ \boldsymbol{q}_4 \right ]$};
			
			\draw [chainLine, ->, mygray, line width=0.8pt] (r1) [] to (m1v3);
			\draw [chainLine, ->, mygray, line width=0.8pt] (v3vv) to (m1v3);
			
			\node[right=of m1v3, fill=darkgrass!40, minimum width=15mm, minimum height=6mm] (op2) [] {FFN$_{op=\text{``}\times\text{''}}$};
			\draw [chainLine, ->, mygray, line width=0.8pt] (m1v3) to (op2);
			
			\node[expr, right =of op2, minimum width=15mm, minimum height=6mm, xshift=5mm] (m2) [] {$\boldsymbol{e}_{3,4,\times}$};
			%, label=above:{$e_{3,4,*}^{(2)}$}
			\draw [chainLine, ->, mygray, line width=0.8pt] (op2) to (m2);

			\begin{pgfonlayer}{background}
				\draw [chainLine, dashed, mygray, line width=0.6pt] (v1v) [out=-120,in=120, looseness=1]  to (v1vv);
				\draw [chainLine, dashed, mygray, line width=0.6pt] (v2v) [out=-120,in=120, looseness=1] to (v2vv);
				\draw [chainLine, dashed, mygray, line width=0.6pt] (v3v)  [out=-60,in=30, looseness=1]to (v3vv);
				\draw [chainLine, dashed, mygray, line width=0.6pt] (m1)  [out=-120,in=120, looseness=1]to (r1);
			\end{pgfonlayer}
			
		\end{tikzpicture}
	}
	\vspace{-2mm}
	\caption{Model architecture for the deductive reasoner. We show the inference procedure to obtain the expression  ``$q_1 \div q_2 \times q_3$'' for the example question.}
	\label{fig:model}
\end{figure}
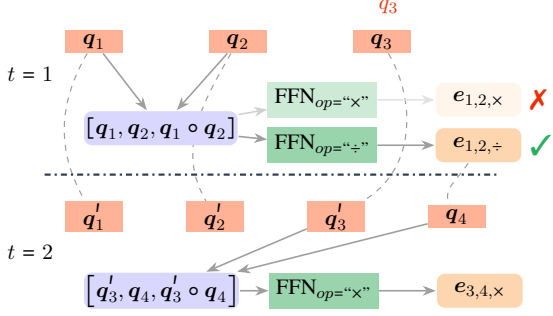

\paragraph{Reasoner} 
Figure \ref{fig:model} shows the deductive reasoning procedure in our model for an example that involves $3$ quantities.
We first convert the quantities (e.g., \textit{$2,088$}) into a general quantity token ``\textit{<quant>}''.
We next adopt a pre-trained language model such as BERT~\cite{devlin2019bert} or Roberta \cite{cui2019pre,liu2019roberta} to obtain the quantity representation $\boldsymbol{q}$ for each quantity $q$. 
%In terms of the constant, we simply use constant embedding $\mathbf{C}^{\mathcal{C} \times d }$ to represent each constant in $\mathcal{C}$.  
Given the quantity representations, we consider all the possible quantity pairs, $(q_i, q_j)$.
Similar to \citet{lee2017end}, we can obtain the representation of each pair by concatenating the two quantity representations and the element-wise product between them. 
As shown in Figure \ref{fig:model}, we apply a non-linear feed-forward network (FFN) on top of the pair representation to get the  representation of the newly created expression. 
The above procedure can be mathematically written as:
% \begin{center}
	% 			\begin{tabular}{c}
		% 					$
		% 					\boldsymbol{e}_{i,j, op} ={\ffn}_{op} (\left [\boldsymbol{q}_i, \boldsymbol{q}_j, \boldsymbol{q}_i \circ \boldsymbol{q}_j \right ]),  ~~ i\leq j
		% 					$
		% 				\end{tabular}
	% 			%\vspace{-2mm}
	% 		\end{center}
% \vspace{-1mm}
\begin{equation}
	\begin{split}
		\boldsymbol{e}_{i,j, op} ={\ffn}_{op} (\left [\boldsymbol{q}_i, \boldsymbol{q}_j, \boldsymbol{q}_i \circ \boldsymbol{q}_j \right ]),  ~~ i\leq j
	\end{split}
\end{equation}
where $\boldsymbol{e}_{i,j,op}$ is the representation of the intermediate expression $e$ and $op$ is the operation  (e.g., ``$+$'', ``$-$'') applied to the ordered pair ($q_i$, $q_j$). 
$\ffn_{op}$ is an operation-specific network that gives the expression representation under the particular operation $op$.
%To obtain the representation from all possible pairs, we make a constraint $i \leq j $ to ensure we do not repeat the computation. 
Note that we have the constraint $i \leq j $.
As a result we also consider the ``\textit{reverse operation}'' for division and subtraction \cite{roy2015solving}.

As shown in Figure \ref{fig:model}, the expression $\boldsymbol{e}_{1,2,\div}$ will be regarded as a new quantity with representation $\boldsymbol{q}_4$ at $t=1$.  
In general, we can assign a score to a single reasoning step that yields the expression $e_{i,j,op}^{(t)}$ from $q_i$ and $q_j$ with operation $op$.
Such a score can be calculated by summing over the scores defined over the representations of the two quantities and the score defined over the expression:
\begin{equation}
	% 	\begin{split}
		s(e_{i,j,op}^{(t)})  = s_q(\boldsymbol{q}_i) + s_q(\boldsymbol{q}_j) + s_e (\boldsymbol{e}_{i,j, op})
		% 		s_q(\boldsymbol{q}_i) &= \vec{w}_q \cdot \ffn
		% 		(\boldsymbol{q}_i) \\
		% 		s (\boldsymbol{e}_{i,j, op}) & = \vec{w}_e \cdot \boldsymbol{e}_{i,j, op}
		% 	\end{split} 
\end{equation}
where we have:
\begin{equation}
	\begin{split}
		s_q(\boldsymbol{q}_i) &= \vec{w}_q \cdot \ffn
		(\boldsymbol{q}_i) \\
		s_e (\boldsymbol{e}_{i,j, op}) & = \vec{w}_e \cdot \boldsymbol{e}_{i,j, op}
	\end{split}
\end{equation}
where $s_q(\cdot)$ and $s_e(\cdot)$ are the scores assigned to the quantity and the expression, respectively, and
$\vec{w}_q$ and $\vec{w}_e$ are the corresponding learnable parameters. 
Our goal is to find the optimal expression sequence $[e^{(1)}, e^{(2)}, \cdots, e^{(T)}]$ that enables us to compute the final numerical answer, where $T$ is the total number of steps required for this deductive process.

\begin{table}[t!]
	\centering
	\resizebox{1.0\linewidth}{!}{
		\begin{tabular}{lc}
			\toprule
			%% whether we add the duplication number or not
			\textbf{Rationalizer} & \textbf{Mechanism}  \\
			\midrule
			Multi-head Self-Attention & $\text{Attention}(Q=\left [\boldsymbol{q}_i, \boldsymbol{e}\right ], K=\left [ \boldsymbol{q}_i, \boldsymbol{e} \right ], V=\left [ \boldsymbol{q}_i, \boldsymbol{e} \right ])$ \\
			GRU cell & $\text{GRU\_Cell}(\text{input} = \boldsymbol{q}_i, \text{previous hidden} = \boldsymbol{e})$ \\
			\bottomrule
		\end{tabular}
	}
	\vspace{-1mm}
	\caption{The mechanism in different rationalizers.}
	\label{tab:diff_rationalizer}
\end{table}

\paragraph{Terminator} 
Our model also has a mechanism that decides whether the deductive procedure is ready to terminate at any given time.
We introduce a binary label $\tau$, where $1$ means the procedure stops here, and $0$ otherwise.
The final score of the expression $e$ at time step $t$ can be  calculated as:
% \begin{center}
	% 		\begin{tabular}{c}
		% 			$
		% 			s\big (e_{i,j,op}^{(t)}, \tau \big ) = s(e_{i,j,op}^{(t)}) + \vec{w}_{\tau} \cdot \ffn (\boldsymbol{e}_{i,j,op})
		% 			$
		% 		\end{tabular}
	% 		%\vspace{-2mm}
	% 	\end{center}

\begin{equation}
	S\big (e_{i,j,op}^{(t)}, \tau \big ) = s(e_{i,j,op}^{(t)}) + \vec{w}_{\tau} \cdot \ffn (\boldsymbol{e}_{i,j,op})
\end{equation}
where $\vec{w}_{\tau}$ is the parameter vector for scoring the  $\tau$. 

\definecolor{keppel}{RGB}{88, 177, 159}

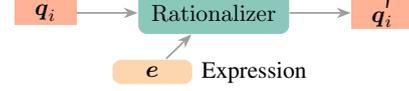
\begin{figure}[t!]
	\centering
	\adjustbox{max width=0.7\linewidth}{
		\begin{tikzpicture}[node distance=1.0mm and 5.0mm, >=Stealth, 
			wordnode/.style={draw=none, minimum height=5mm, inner sep=0pt},
			chainLine/.style={line width=1.1pt,-, color=deepblue},
			entbox/.style={draw=black, rounded corners, fill=red!20, dashed},
			mathop/.style={draw=deepblue, circle, minimum height=7mm, inner sep=1pt, line width=1pt},
			quant/.style={draw=deeppurple, circle, minimum height=7mm, inner sep=1pt, line width=1pt},
			expr/.style={draw=none, rectangle, minimum width=10mm, inner sep=1pt, line width=1pt, rounded corners, minimum height=4mm, fill=puffin!40},
			invis/.style={draw=none},
			plm/.style={draw=none, fill=lowblue!60, rectangle, rounded corners, minimum width=140mm, minimum height=9mm},
			score/.style={draw=none, fill=myred!40, circle, minimum width=5mm},
			vector/.style={draw=none, fill=coral!60, minimum height=4mm, minimum width=10mm, inner sep=1pt}
			]

			\node[vector] (vi)[]{$\boldsymbol{q}_i$}; 
			
			\node[right=of vi, fill=keppel!70, minimum width=15mm, minimum height=6mm, inner sep=6pt, rounded corners, xshift=5mm] (rat) [] {$\rationalizer$};
			
			\node[vector, right=of rat, xshift=5mm] (viprime)[]{$\boldsymbol{q}_i^{\prime}$};

			\node[expr, below =of rat, minimum width=13mm, minimum height=4mm, xshift=-10mm, yshift=-3mm, label=right:{Expression}] (m1) [] {$\boldsymbol{e}$};
			
			\draw [chainLine, mygray, line width=0.8pt, ->] (vi)  []to (rat);
			\draw [chainLine, mygray, line width=0.8pt, ->] (m1)  []to (rat);
			\draw [chainLine, mygray, line width=0.8pt, ->] (rat)  []to (viprime);
			
		\end{tikzpicture}
	}
	\vspace{-1mm}
	\caption{Rationalizing quantity representation.}
	\label{fig:rationalizer}
\end{figure}

\paragraph{Rationalizer} 

Once we obtain a new intermediate expression at step $t$, it is crucial to update the representations for the existing quantities.
We call this step {\em rationalization} because it could potentially give us the rationale that explains an outcome~\cite{lei2016rationalizing}.
As shown in Figure \ref{fig:rationalizer}, the intermediate expression $\boldsymbol{e}$ serves as the rationale that explains how the quantity changes from $\boldsymbol{q}$ to $\boldsymbol{q}^{\prime}$.
Without this step, there is a potential shortcoming for the model.
That is, because if the quantity representations do not get updated as we continue the deductive reasoning process,
those expressions that were initially highly ranked (say, at the first step) would always be preferred over those lowly ranked ones
throughout the process.\footnote{See the supplementary material for more details on this.}
%to always choose the previous intermediate expression $e_{t}$ as the best expression for next step $t+1$ (especially when $t$ is large, such as $t>=3$).
%For example, the first two steps in solving the expression $(3+4)*(5+6)$ are $3+4$ and $5+6$.
%The quantity representations for these four quantities are never updated.
%After we obtain $e^{(1)} = 3+4$, we will never obtain the $5+6$ as $e^{(2)}$.
%Because the score for $3+4$ is always better than $5+6$. 
%Our empirical experiments in Table \ref{tab:rationalizer} further demonstrate this phenomenon.
We rationalize the quantity representation using the current intermediate expression  $e^{(t)}$, so that the quantity is aware of the generated expressions when its representation gets updated.
This procedure can be mathematically formulated as follows:
\begin{equation}
	\boldsymbol{q}_{i}^{\prime} = \rationalizer(\boldsymbol{q}_{i}, \boldsymbol{e}^{(t)}) ~~~  \forall~ 1  \leq i \leq  \vert \mathcal{Q} \vert 
	\label{equ:rationalizer}
\end{equation}
%In Equation \ref{equ:rationalizer}, t

Two well-known techniques we can adopt as rationalizers are 
\textit{multi-head self-attention}~\cite{vaswani2017attention} and a \textit{gated recurrent unit} (GRU)~\cite{cho2014properties} cell, which allow us to update the quantity representation, given the intermediate expression representation.
Table \ref{tab:diff_rationalizer} shows the mechanism in two different rationalizers. 
%In self-attention, we essentially first
For the first approach, we essentially construct a sentence with two token representations -- quantity $\boldsymbol{q}_i$ and the previous expression $\boldsymbol{e}$ -- to perform self-attention.
%are the {key} and {value}\footnote{\textcolor{darkgrass}{The combined representations $\left [\boldsymbol{q}_i, \boldsymbol{e} \right ]$ of quantity and expression can be seen as a sentence with two words.}}. 
\textcolor{black}{In the second approach, we use $\boldsymbol{q}_i$ as the {input state} and $\boldsymbol{e}$ as the previous hidden state in a GRU cell.}

\begin{table}[t!]
	\centering
	\resizebox{\linewidth}{!}{
		\begin{tabular}{lcccccc}
			\toprule
			%% whether we add the duplication number or not
			% 			\textbf{Dataset} & \#\textbf{Train} & \#\textbf{Valid} & \#\textbf{Test} & \textbf{Avg. Sent\_len} & \textbf{\#Const.} \\
			\multirow{2}{*}{\textbf{Dataset}} & \multirow{2}{*}{\#\textbf{Train}} & \multirow{2}{*}{\#\textbf{Valid}} & \multirow{2}{*}{\#\textbf{Test}} & \textbf{Avg.} & \multirow{2}{*}{\textbf{\#Const.}} & \multirow{2}{*}{\textbf{Lang.}}\\
			& & & & \textbf{Sent Len} & & \\
			\midrule
			MAWPS & \textcolor{white}{0}1,589 & \textcolor{white}{0,}199 & \textcolor{white}{0,}199 &  30.3 & 17 &English\\ 
			Math23k & 21,162 & 1,000 & 1,000 &  26.6 & {\color{white}0}2 & Chinese\\
			MathQA$\dagger$ & 16,191 & 2,411 & 1,605 & 39.6 & 24 & English\\
			SVAMP &  \textcolor{white}{0}3,138 & - & 1,000& 34.7 & 17 & English\\
			\bottomrule
		\end{tabular}
	}
	\vspace{-1mm}
	\caption{Dataset statistics. $\dagger$: we follow \citet{tan2021investigating} to do preprocessing and obtain the subset.}
	\label{tab:datastat}
\end{table}

\subsection{Training and Inference}\label{sec:train}
Similar to training sequence-to-sequence models~\cite{luong2015effective}, we adopt the teacher-forcing strategy~\cite{williams1989learning} to guide the model with gold expressions during training. 
The loss\footnote{Actually, one might have noticed that this loss comes with a trivial solution at ${\boldsymbol\theta}=\boldsymbol{0}$. In practice, however, our model and training process would prevent us from reaching such a degenerate solution with proper initialization~\cite{goodfellow2016deep}. This is similar to the training of a structured perceptron \cite{collins2002discriminative}, where a similar situation is also involved.} can be written as:
\begin{equation}
	\begin{split}
		\mathcal{L}(\boldsymbol{\theta})   =& \sum_{t=1}^{T} \Big ( \max_{(i,j,op)\in\mathcal{H}^{(t)}, \tau} \Big[ \mathcal{S}_{\boldsymbol{\theta}} (e_{i,j, op}^{(t)}, \tau) \Big] \\
		& - \mathcal{S}_{\boldsymbol{\theta}} (e_{i^*,j^*, op^*}^{(t)}, \tau^*) \Big ) + \lambda ||\boldsymbol{\theta} ||^2
	\end{split}
	\label{equ:loss}
\end{equation}
% \ell_{t} (\boldsymbol{\theta}) 
where $\boldsymbol{\theta}$ includes all parameters in the deductive reasoner and $\mathcal{H}^{(t)}$ contains all the possible choices of quantity pairs and relations available at time step $t$. 
$\lambda$ is the hyperparameter for the $L_2$ regularization term.
The set $\mathcal{H}^{(t)}$ grows as new expressions are constructed and become new quantities during the deductive reasoning process.
The overall loss is computed by summing over the loss at each time step (assuming totally $T$ steps).

% \begin{center}
	% 		\begin{tabular}{c}
		% 			$
		% 			\mathcal{L}(\boldsymbol{\theta})  = \sum_{t=1}^{T} \ell_t(\boldsymbol{\theta}) 
		% 			$
		% 		\end{tabular}
	% 		%\vspace{-2mm}
	% 	\end{center}
% \begin{equation}
	% 	\mathcal{L}(\boldsymbol{\theta})  = \sum_{t=1}^{T} \ell_t(\boldsymbol{\theta}) 
	% 	\label{equ:overall_loss}
	% \end{equation}
During inference, we set a maximum time step $T_{max}$ and find the best expression $e^*$ that has the highest score at each time step. 
Once we see $\tau = 1$ is chosen, we stop constructing new expressions and terminate the process.
The overall expression (formed by the resulting expression sequence) will be used for computing the final numerical answer.

\paragraph{Declarative Constraints} 

Our model repeatedly relies on existing quantities to construct new quantities, which results in a structure showing the deductive reasoning process.
% \textcolor{black}{The procedure is analogous to that of classic constituency parsing~\cite{kay1980algorithm} where we design context-free grammars~\cite{booth1969probabilistic} to constrain the structured space.}
One advantage of such an approach is that it allows certain declarative knowledge to be conveniently incorporated. 
% {\color{blue}While it is also possible for other models such as sequence-to-sequence based models to incorporate certain declarative knowledge~\cite{hokamp2017lexically}, we believe our model has the additional convenience in certain scenarios.}
For example, as we can see in Equation \ref{equ:loss}, the default approach considers all the possible combinations among the quantities during the maximization step.
We can easily impose constraints to avoid considering certain combinations.
In practice, we found in certain datasets such as SVAMP, there does not exist any expression that involve operations applied to the same quantity (such as $9+9$ or $9\times 9$, where $9$ is from the same quantity in the text).
Besides, we also observe that the intermediate results would not be negative. 
We can simply exclude such cases in the maximization process, effectively reducing the search space during both training and inference.
We show  that adding such declarative constraints can help improve the performance.
% \footnote{While other declarative knowledge, such as prior world knowledge (e.g., ``there are 31 days in March'') could also be incorporated into our model, we did not exploit them in this work so as to ensure fair comparisons with previous efforts. We leave them for future explorations.}

%\footnote{In addition, our model also allows us to impose constraints to disable certain operations for certain combinations of quantities. For example, some unary operators cannot be considered for two quantities though these cases are not considered in this work.{\color{red}not clear what you mean by the last sentence. Do you think you can say things such as 1 year = 365 days March as 31 days etc as a declarative knowledge that can be added?}}

\section{Experiments}
\label{sec:experiments}

\begin{figure}[t!]
	\pgfplotsset{every axis/.append style={
			%		axis x line=middle,    % put the x axis in the middle
			%		axis y line=middle,    % put the y axis in the middle
			%		axis line style={<->}, % arrows on the axis
			%		xlabel={$x$},          % default put x on x-axis
			%		ylabel={$y$},          % default put y on y-axis
			label style={font=\footnotesize},
			tick label style={font=\footnotesize}  
	}}
	\pgfplotstableread[row sep=\\,col sep=&]{
		interval & mawps & math23k &mathqa &  svamp \\
		$1$     & 60.76  & 19.83  & 7.03 & 38.15  \\
		$2$     & 37.38 & 49.99  & 15.82 & 29.4 \\
		$3$    & 1.56 & 20.07 & 21.55 & 24.45 \\
		$4$   & 0.25 & 6.29 & 16.76 & 7.65\\
		$5$   & 0.05  & 2.45 & 13.87 & 0.15 \\
		$~~\geq 6$     & 0. & 1.35 & 24.98 & 0.\\
	}\mydata
	\centering
	\begin{subfigure}[b]{0.2\textwidth}
		\centering
		\adjustbox{max width=0.9\textwidth}{
			\begin{tikzpicture}
				\begin{axis}[
					axis line style = thick,
					ybar,bar width=9,legend style={nodes={scale=0.3, transform shape}, line width=0.2mm}, height=3cm,
					symbolic x coords={$1$, $2$, $3$, $4$, $5$, $~~\geq 6$ },
					xtick=data, ymin = 0, ymax = 70, line width = 0.5mm]
					\addplot [pattern = horizontal lines, pattern color = blue, draw=blue] table[x=interval,y=mawps]{\mydata};
					\legend {\LARGE MAWPS};
				\end{axis}
			\end{tikzpicture}
		}
	\end{subfigure}
	\begin{subfigure}[b]{0.2\textwidth}
		\centering
		\adjustbox{max width=0.9\textwidth}{
			\begin{tikzpicture}
				\begin{axis}[
					axis line style = thick,
					ybar,bar width=9,legend style={nodes={scale=0.25, transform shape}, line width=0.2mm}, height=3cm,
					symbolic x coords={$1$, $2$, $3$, $4$, $5$, $~~\geq 6$ },
					xtick=data, ymin = 0, ymax = 70, line width = 0.5mm]
					\addplot [pattern = grid, pattern color = teal, draw=teal] table[x=interval,y=math23k]{\mydata};
					\legend {\LARGE Math23k};
				\end{axis}
			\end{tikzpicture}
		}
	\end{subfigure}
	\\
	\begin{subfigure}[b]{0.2\textwidth}
		\centering
		\adjustbox{max width=0.9\textwidth}{
			\begin{tikzpicture}
				\begin{axis}[
					axis line style = thick,
					ybar,bar width=9,legend style={nodes={scale=0.3, transform shape}, line width=0.2mm}, height=3cm,
					symbolic x coords={$1$, $2$, $3$, $4$, $5$, $~~\geq 6$ },
					xtick=data, ymin = 0, ymax = 60, line width = 0.5mm]
					\addplot [pattern = dots, pattern color = violet, draw=violet] table[x=interval,y=mathqa,]{\mydata};
					\legend {\LARGE MathQA};
				\end{axis}
			\end{tikzpicture}
		}
	\end{subfigure}
	\begin{subfigure}[b]{0.2\textwidth}
		\centering
		\adjustbox{max width=0.9\textwidth}{
			\begin{tikzpicture}
				\begin{axis}[
					axis line style = thick,
					ybar,bar width=9, legend style={nodes={scale=0.3, transform shape}, line width=0.2mm}, height=3cm,
					symbolic x coords={$1$, $2$, $3$, $4$, $5$, $~~\geq 6$ },
					xtick=data, ymin = 0, ymax = 60, line width = 0.5mm]
					\addplot [pattern = north east lines, pattern color = orange, draw=orange] table[x=interval,y=svamp,]{\mydata};
					\legend{\Large SVAMP};
				\end{axis}
			\end{tikzpicture}
		}
	\end{subfigure}
	\vspace{-3mm}
	\caption{Percentage of questions with different operation count.}
	\label{fig:data_opnum}
\end{figure}
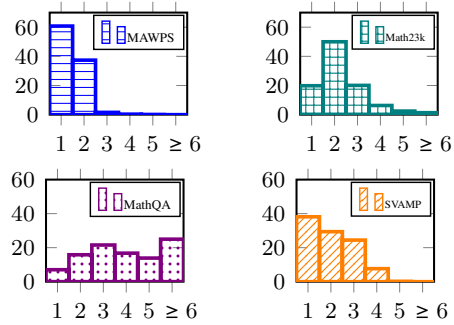

\paragraph{Datasets}
\label{sec:dataset}

We conduct experiments on four datasets across two different languages: MAWPS~\cite{koncel2016mawps}, Math23k~\cite{wang2017deep}, MathQA~\cite{amini2019mathqa}, and SVAMP~\cite{patel2021are}.
% MAWPS, MathQA, and SVAMP are in English. Math23k is in Chinese.
The dataset statistics can be found in Table \ref{tab:datastat}. 
For MathQA\footnote{The original MathQA~\cite{amini2019mathqa} dataset contains a certain number of instances that have annotated equations which cannot lead to the correct numerical answer.}, we follow \citet{tan2021investigating}\footnote{Our dataset size is not exactly the same as \citet{tan2021investigating} as they included some instances that are wrongly annotated. We only kept the part that has correct annotations. We confirmed such information with the authors of \citet{tan2021investigating}, and make our version of this dataset publicly available.} to adapt the dataset to filter out some questions that are unsolvable. 
We consider the operations ``\textit{addition}'', ``\textit{subtraction}'', ``\textit{multiplication}'', and ``\textit{division}'' for MAWPS and SVAMP, and an extra ``\textit{exponentiation}'' for MathQA and Math23k.
% An extra ``\textit{exponentiation}'' operation is considered for Math23k and MathQA.
%We consider all the constants that appear in the training set in our constant embedding $\mathbf{C}$.  
%Following previous work on MAWPS~\cite{lan2021mwptoolkit,zhang2020graph}, we report the performance of 5-fold cross-validation performance.
%Previous works reports different setting in evaluating Math23k, we follow the common practice~\cite{xie2019goal,li2019modeling,zhang2020graph} and report both the 5-fold cross-validation  and test set performance.

The number of operations involved in each question can be one of the indicators to help us gauge the difficulty of a dataset.
Figure \ref{fig:data_opnum} shows the percentage distribution of the number of operations involved in each question.
The MathQA dataset generally contains larger portions of questions that involve more operations, while 97\% of the questions in MAWPS can be answered with only one or two operations. 
More than 60\% of the instances in MathQA have three or more operations, which likely makes their problems harder to solve.
% as compared to MAWPS and Math23k.
Furthermore, MathQA~\cite{amini2019mathqa} contains GRE questions in many domains including physics, geometry, probability, etc., while Math23k questions are from primary school.
Different from other datasets, SVAMP~\cite{patel2021are}\footnote{There is no test split for this dataset. We strictly follow the experiment setting in \citet{patel2021are}.} is a challenging set that is manually created to evaluate a model's robustness. 
They applied variations over the instances sampled from MAWPS.
Such variations could be: adding extra quantities, swapping the positions between noun phrases, etc.

% \textcolor{blue}{}
% {\color{red}not very clear. can you re-write this sentence? They make modifications on the questions based on the errors made by existing models~\cite{zhang2020graph}.}

%\citet{patel2021are} created the challenging SVAMP dataset with 1,000 test instances to evaluate the robustness of a model. 

%Following previous work, we report both equation accuracy and solution accuracy. 

% \paragraph{Expression Duplication}
% As mentioned in Section \ref{sec:intro}, it is intuitive reuse the expression that are previously computed. 
% We found that there is 1\% of the annotations in Math23k have the expression duplication issue, where the same expression appear more than twice in the annotation. 
% During data preprocess, we remove those duplication in our training data. 

\begin{table}[t!]
	\centering
	\setlength{\tabcolsep}{2pt} % Default value: 6pt
	\renewcommand{\arraystretch}{1.1} % Default value: 1
	\resizebox{0.9\linewidth}{!}{
		\begin{tabular}{p{0.05\textwidth}lc}
			\toprule
			\multicolumn{2}{c}{\textbf{Model}} & \multicolumn{1}{c}{\textbf{Val Acc.}} \\
			\midrule
			%\midrule
			\parbox[t]{2mm}{\multirow{4}{*}{\rotatebox[origin=c]{90}{\textsc{S2S}}}} & GroupAttn~\cite{li2019modeling} &  76.1 \\
			& Transformer~\cite{vaswani2017attention}  & 85.6 \\
			%		& mBERT+LSTM & & \\
			& BERT-BERT~\cite{lan2021mwptoolkit} &  86.9  \\
			& Roberta-Roberta~\cite{lan2021mwptoolkit} & 88.4  \\
			%			& mBART~\cite{shen2021generate} & 84.0\\
			%			& Gen\&Rank~\cite{shen2021generate} & 84.0\\
			\midrule
			
			\parbox[t]{2mm}{\multirow{4}{*}{\rotatebox[origin=c]{90}{\textsc{S2T/G2T}}}}	& GTS~\cite{xie2019goal} &  82.6  \\
			& Graph2Tree~\cite{zhang2020graph}& 85.6  \\
			%		Seq2tree/ & BERT+GTS (no CL) & & \\
			%		Graph2tree & BERT + GTS (with CL) & & \\
			& Roberta-GTS~\cite{patel2021are}&  88.5  \\
			& Roberta-Graph2Tree~\cite{patel2021are}& 88.7  \\
			\midrule
			\parbox[t]{2mm}{\multirow{4}{*}{\rotatebox[origin=c]{90}{\textsc{Ours}}}} & \textsc{Bert-DeductReasoner} & 91.2 {\small{($\pm$ 0.16)}}\\
			& \textsc{Roberta-DeductReasoner}& \textbf{92.0} {\small{($\pm$ 0.20)}}\\
			& \textsc{mBERT-DeductReasoner}   &   91.6 {\small{($\pm$ 0.13)}}\\
			& \textsc{XLM-R-DeductReasoner} &  91.6 {\small{($\pm$ 0.11)}}\\
			\bottomrule
		\end{tabular}
	}
	\vspace{-2mm}
	\caption{5-fold cross-validation results on MAWPS. }
	\label{tab:mawpsres}
\end{table}

\begin{table}[t!]
	\centering
	\setlength{\tabcolsep}{3pt} % Default value: 6pt
	\renewcommand{\arraystretch}{1.1} % Default value: 1
	\resizebox{0.95\linewidth}{!}{
		\begin{tabular}{p{0.05\textwidth}lcc}
			\toprule
			\multicolumn{2}{c}{\multirow{2}{*}{\textbf{Model}}} & \multicolumn{2}{c}{\textbf{Val Acc.}} \\
			& & \textbf{Test} & \textbf{5-fold} \\
			\midrule
			\parbox[t]{2mm}{\multirow{4}{*}{\rotatebox[origin=c]{90}{\textsc{S2S}}}}& GroupAttn~\cite{li2019modeling} & 69.5 & 66.9 \\
			& mBERT-LSTM~\cite{tan2021investigating} & 75.1 & - \\
			& BERT-BERT~\cite{lan2021mwptoolkit} & - & 76.6 \\
			& Roberta-Roberta~\cite{lan2021mwptoolkit} & - & 76.9 \\
			%		& mBART$\ddagger$~\cite{shen2021generate} & 80.8 & 80.0\\
			%		& Gen\&Rank$\ddagger$~\cite{shen2021generate} & 85.4 & 84.3 \\
			\midrule
			\parbox[t]{2mm}{\multirow{8}{*}{\rotatebox[origin=c]{90}{\textsc{S2T/G2T}}}}& GTS~\cite{xie2019goal} & 75.6 & 74.3 \\
			& KA-S2T$\dagger$~\cite{wu2020knowledge} & 76.3 & - \\
			& MultiE\&D~\cite{shen2020solving} & 78.4 & 76.9 \\
			& Graph2Tree~\cite{zhang2020graph} & 77.4 & 75.5 \\
			& NeuralSymbolic~\cite{qin2021neural} & - & 75.7 \\
			& NUMS2T$\dagger$~\cite{wu2021math} & 78.1 & - \\
			& HMS~\cite{lin2021hms} & 76.1 &- \\
			& BERT-Tree~\cite{li2021seeking} & 82.4 & - \\
			%		& BERT + GTS (with CL)~\cite{li2021seeking} & 83.2 & - \\
			\midrule
			\parbox[t]{2mm}{\multirow{4}{*}{\rotatebox[origin=c]{90}{\textsc{Ours}}}} & \textsc{Bert-DeductReasoner}  &  84.5   {\small{($\pm$ 0.16)}}& 82.6 {\small{($\pm$ 0.17)}} \\
			& \textsc{Roberta-DeductReasoner} &  \textbf{85.1} {\small{($\pm$ 0.24)}} & \textbf{83.0} {\small{($\pm$ 0.23)}}\\
			& \textsc{mBERT-DeductReasoner}     &  84.3 {\small{($\pm$ 0.19)}} & 82.5 {\small{($\pm$ 0.33)}}\\
			& \textsc{XLM-R-DeductReasoner}   &  84.0 {\small{($\pm$ 0.22)}} &  82.0 {\small{($\pm$ 0.12)}}\\
			\bottomrule
		\end{tabular}
	}
	\vspace{-1mm}
	\caption{Results on Math23k. $\dagger$: they used their own splits {\color{black}(so their results may not be directly comparable)}.}
	\label{tab:math23kres}
\end{table}

\paragraph{Baselines}
% We compare with existing works that reported results on these four datasets. 
The baseline approaches can be broadly categorized into sequence-to-sequence (S2S), sequence-to-tree (S2T) and graph-to-tree (G2T) models.
\textbf{GroupAttn}~\cite{li2019modeling} designed several types of attention mechanisms such as question or quantity related attentions in the seq2seq model.
\citet{tan2021investigating} uses multilingual BERT with an LSTM decoder~(\textbf{mBERT-LSTM}). 
\citet{lan2021mwptoolkit} presented two seq2seq models that use BERT/Roberta as both encoder and decoder, namely, \textbf{BERT-BERT} and \textbf{Roberta-Roberta}. 
% These approaches are mainly categorized into sequence-to-sequence (S2S)~\cite{tan2021investigating,lan2021mwptoolkit} and sequence-to-tree (S2T)~\cite{zhang2020graph,xie2019goal,lan2021mwptoolkit,li2021seeking} models.
% These approaches are mainly categorized into sequence-to-sequence (S2S)~\cite{tan2021investigating,lan2021mwptoolkit} and sequence-to-tree (S2T)~\cite{zhang2020graph,xie2019goal,lan2021mwptoolkit,li2021seeking} models.
% S2S approaches 
Sequence-to-tree models mainly use a tree-based decoder with GRU (\textbf{GTS})~\cite{xie2019goal} or BERT as the encoder~(\textbf{BERT-Tree})~\cite{liang2021mwp,li2021seeking}.
% \textbf{GTS}~\cite{xie2019goal} and BERT with a tree decoder (\textbf{BERT-Tree})~\cite{liang2021mwp,li2021seeking}.
\textbf{NUMS2T}~\cite{wu2020knowledge} and \textbf{NeuralSymbolic}~\cite{qin2021neural} solver incorporate external knowledge in the S2T architectures.
\textbf{Graph2Tree}~\cite{zhang2020graph} models the quantity relations using GCN. 
%Tree-based models mainly use the goal-driven tree structures (\textbf{GTS})~\cite{xie2019goal} as the decoder. 
% Tree-based models mainly include GTS~\cite{xie2019goal}, \textbf{Graph2Tree} (G2T)~\cite{zhang2020graph} and 
%We compare with \textbf{Graph2Tree}~\cite{zhang2020graph} which models the quantities using graph convolutional networks (GCN)~\cite{kipf2017semi}. 
% Other works~\cite{liang2021mwp,li2021seeking} use a pre-trained language model as encoder coupled with the GTS decoder (\textbf{BERT-GTS}).
%\citet{liang2021mwp} presented a strong baseline that use BERT as encoder and GTS as decoder.
%\citet{li2021seeking} presented a same model architecture and proposed to use contrastive leanring~\cite{hadsell2006dimensionality} to learn the difference between different expression patterns.

\paragraph{Training Details}
We adopt BERT~\cite{devlin2019bert} and Roberta~\cite{liu2019roberta} for the English datasets.
Chinese BERT and {\color{black}Chinese} Roberta~\cite{cui2019pre} are used for Math23k.
We use the GRU cell as the rationalizer.
% To further evaluate the multilingual language models, 
We also conduct experiments with  multilingual BERT and XLM-Roberta~\cite{conneau2020unsupervised}.
The pre-trained models are initialized from HuggingFace's Transformers~\cite{wolf2020transformers}.
% All parameters are fine-tuned during training. 
We optimize the loss with the Adam optimizer~\cite{kingma2014adam,loshchilov2018decoupled}.
% in equation \ref{equ:overall_loss}
We use a learning rate of $2e$-$5$ and a batch size of $30$. 
The regularization coefficient $\lambda$ is set to $0.01$.
We run our models with $5$ random seeds and report the average results (with standard deviation).
Following most previous works, we mainly report the value accuracy (percentage) in our experiments. 
In other words, a prediction is considered correct if the predicted expression leads to the same value as the gold expression.
Following previous practice~\cite{zhang2020graph,tan2021investigating,patel2021are}, we report  $5$-fold cross-validation results on both MAWPS\footnote{All previous efforts combine training/dev/test sets and perform $5$-fold cross validation, which we follow.} and Math23k, {\color{black}and also report the test set performance for Math23k, MathQA and SVAMP. }
% Previous works report different settings when evaluating on Math23k. Thus, we follow the common practice~\cite{xie2019goal,li2019modeling,zhang2020graph} and report both the 5-fold cross-validation and test set performance.
%Following previous work, we report both equation accuracy and solution accuracy. 

\begin{table}[t!]
	\centering
	\setlength{\tabcolsep}{2pt} % Default value: 6pt
	\renewcommand{\arraystretch}{1.1} % Default value: 1
	\resizebox{0.76\linewidth}{!}{
		\begin{tabular}{lc}
			\toprule
			\multicolumn{1}{c}{\textbf{Model}} & \textbf{Val Acc.} \\
			\midrule
			Graph2Tree \cite{zhang2020graph} & 69.5 \\
			BERT-Tree~\cite{li2021seeking} & 73.8 \\
			% 			Roberta-Tree~\cite{li2021seeking} & 73.6 \\
			mBERT+LSTM~\cite{tan2021investigating} & 77.1 \\
			\midrule
			\textsc{Bert-DeductReasoner}  &  78.5   {\small{($\pm$ 0.07)}}\\
			\textsc{Roberta-DeductReasoner} &  \textbf{78.6} {\small{($\pm$ 0.09)}} \\
			\textsc{mBERT-DeductReasoner}     &  78.2 {\small{($\pm$ 0.21)}}\\
			\textsc{XLM-R-DeductReasoner}   &  78.2 {\small{($\pm$ 0.11)}}\\
			\bottomrule
		\end{tabular}
	}
	\vspace{-2mm}
	\caption{Test accuracy comparison on MathQA.}
	\label{tab:mathqares}
\end{table}

\subsection{Results}

\paragraph{MAWPS and Math23k}

We first discuss the results on MAWPS and Math23k, two datasets that are commonly used in previous research.
% The former is in English and is a smaller dataset, while the latter is a larger dataset in Chinese.
Table \ref{tab:mawpsres} and \ref{tab:math23kres} show the main results of the proposed models with different pre-trained language models.
We compare with previous works that have reported results on these datasets.
Among all the encoders for our model \textsc{DeductReasoner}, the Roberta encoder achieves the best performance.
In addition, \textsc{DeductReasoner} significantly outperforms all the baselines regardless of the choice of encoder.
The performance on the best S2S model (Roberta-Roberta) is on par with the {\color{black}best} S2T model (Roberta-Graph2Tree) on MAWPS.
% {\color{brown}However, S2S models generally perform worse than the S2T/G2T models on Math23k.
	% Such results suggest MAWPS is easier to solve as compared to Math23k.}
{\color{black}Overall, the accuracy of Roberta-based \textsc{DeductReasoner} is more than 3 points higher than Roberta-Graph2Tree ($p<0.001$)\footnote{We conduct bootstrapping t-test to compare the results.} on MAWPS, and more than 2 points higher than BERT-Tree ($p<0.005$) on Math23k.}
The comparisons show that our deductive reasoner is robust across different languages and {\color{black}datasets of different sizes}.
We noted that different approaches compare against each other though the experiments are conducted in different settings on the Math23k dataset. 
We present the detailed comparsion in Appendix \ref{sec:math23k_detailed}.

\paragraph{MathQA and SVAMP}

%Next we discuss the results on MathQA and SVAMP, which are more difficult -- the former consists of more complex questions and the latter consists of specifically designed challenging questions.
As mentioned before, MathQA and SVAMP are more challenging -- the former consists of more complex questions and the latter consists of specifically designed challenging questions.
Table \ref{tab:mathqares} and \ref{tab:svampres} show the performance comparisons.
%  with existing works
% As we can see, the absolute accuracy scores are relatively lower as compared with MAWPS and Math23k.
We are able to outperform the best baseline mBERT-LSTM\footnote{We ran the their code on our adapted MathQA dataset.} by 1.5 points in accuracy {\color{black}on MathQA}.
Different from other three datasets, the performance between different language models shows larger gaps on SVAMP. 
As we can see from baselines and our models, the choice of encoder appear to be important for solving questions in SVAMP -- the results on using Roberta as the encoder are particularly striking. 
Our best variant \textsc{Roberta-DeductReasoner} achieves an accuracy score of 47.3 and is able to outperfrom the best baseline (Roberta-Graph2Tree) by 3.5 points ($p<0.01$).
%using Roberta as encoder can yield much better performance than the BERT-based models.
By incorporating the constraints from our prior knowledge (as discussed in \Section\ref{sec:train}), we observe significant improvements for all variants -- up to $7.0$ points for our \textsc{Bert-DeductReasoner}.

% Such improvements indicate the declarative constraints are helpful 
Overall, these results show that our model is more robust as compared to previous approaches on such challenging datasets.
%we are able to further improvement our performance by 1 point.
%For example, both \textsc{Roberta-DeductReasoner} with constraint and Roberta-GTS perform about 10 points better than the \textsc{BERT-DeductReasoner} and BERT-Tree, respectively.
%On the other hand, both multilingual model by \textsc{mBERT} and \textsc{XLM-R} cannot outperform the \textsc{Roberta-DeductReasoner}.
%Such a comparison shows that the monolingual Roberta has a better language understand on the challenging SVAMP.
% \citet{patel2021are} applied certain types of variations (e.g., adding irrelevant information and changing information) on this dataset to make it more challenging compared with MAWPS. 
% Though our deductive reasoner is able to achieve the best performance ($p<0.01$ compared to Roberta-Graph2Tree), the actual number is still lower compared with other datasets. 

\begin{table}[t!]
	\centering
	\setlength{\tabcolsep}{2pt} % Default value: 6pt
	\renewcommand{\arraystretch}{1.1} % Default value: 1
	\resizebox{0.85\linewidth}{!}{
		\begin{tabular}{p{0.05\textwidth}lc}
			\toprule
			\multicolumn{2}{c}{\textbf{Model}} & \textbf{Val Acc.} \\
			\midrule 
			\parbox[t]{2mm}{\multirow{3}{*}{\rotatebox[origin=c]{90}{\textsc{S2S}}}}& GroupAttn~\cite{li2019modeling} & 21.5\\
			&BERT-BERT~\cite{lan2021mwptoolkit}& 24.8 \\
			&Roberta-Roberta~\cite{lan2021mwptoolkit} &30.3\\
			\midrule
			\parbox[t]{2mm}{\multirow{5}{*}{\rotatebox[origin=c]{90}{\textsc{S2T/G2T}}}}& GTS$^{*}$~\cite{xie2019goal} & 30.8 \\
			&Graph2Tree~\cite{zhang2020graph} & 36.5 \\
			&BERT-Tree~\cite{li2021seeking} & 32.4 \\
			%			&Roberta-Tree~\cite{li2021seeking} &29.8 \\
			&Roberta-GTS~\cite{patel2021are} & 41.0 \\
			&Roberta-Graph2Tree~\cite{patel2021are} & 43.8 \\
			\midrule 
			\parbox[t]{2mm}{\multirow{8}{*}{\rotatebox[origin=c]{90}{\textsc{Ours}}}}&\textsc{Bert-DeductReasoner}  &    35.3 {\small{($\pm$ 0.04)}}\\
			&  ~~~~~~+~~\textit{constraints}&42.3 {\small{($\pm$ 0.09)}}\\
			&\textsc{Roberta-DeductReasoner} &  45.0  {\small{($\pm$ 0.10)}}\\
			&  ~~~~~~+~~\textit{constraints}&  \textbf{47.3} {\small{($\pm$ 0.20)}} \\
			&\textsc{mBERT-DeductReasoner}    & 36.1 {\small{($\pm$ 0.07)}} \\
			&  ~~~~~~+~~\textit{constraints}&  41.3   {\small{($\pm$ 0.08)}}\\
			&\textsc{XLM-R-DeductReasoner} &  38.1   {\small{($\pm$ 0.08)}}\\
			&  ~~~~~~+~~\textit{constraints}& 44.6  {\small{($\pm$ 0.15)}}\\
			\midrule
			\midrule
			\multicolumn{3}{l}{\textbf{Additional Experiments} {\footnotesize\it \textcolor{blue}{(experiments conducted after ACL conference)}}} \\
			\multicolumn{3}{l}{{All experiments are incorporated with constraints}} \\
			\multicolumn{2}{l}{\textsc{Roberta-DeductReasoner}$\dagger$} & 48.9 \\
			\multicolumn{2}{l}{\textsc{Deberta-base-DeductReasoner}} & 55.6 \\
			\multicolumn{2}{l}{\textsc{Deberta-v3-Large-DeductReasoner}} & 62.0 \\
			\multicolumn{2}{l}{\textsc{Deberta-v2xx-Large-DeductReasoner}} & 63.6 \\
			\bottomrule
		\end{tabular}
	}
	\vspace{-1.5mm}
	\caption{Test accuracy comparison on SVAMP. $\dagger$: the number is different because we also allow that the pair of quantities in an expression can be the same quantity for all additional experiments.}
	\label{tab:svampres}
\end{table}

\paragraph{Fine-grained Analysis}

{\color{black}We further perform fine-grained performance analysis based on questions with different numbers of operations.}
% As most of the questions (97\%) in MAWPS involve only one or two operations to in the equations, it is generally easier to solve
% As mentioned our assumption in Section \ref{sec:dataset} (Figure \ref{fig:data_opnum}), MAWPS is generally easier to solve while MathQA is harder. 
Table \ref{tab:diffops} shows the accuracy scores for questions that involve different numbers of operations. It also shows the equation accuracy on all datasets\footnote{Equ Acc: we regard an equation as correct if and only if it matches with the reference equation (up to reordering of sub-expressions due to commutative operations, namley ``$+$'' and ``$\times$'').}.
We compared our \textsc{Roberta-DeductReasoner} with the best performing baselines in Table \ref{tab:mawpsres} (Roberta-Graph2Tree), \ref{tab:math23kres} (BERT-Tree), \ref{tab:mathqares} (mBERT+LSTM) and \ref{tab:svampres} (Roberta-Graph2Tree).  
On MAWPS and Math23k, our \textsc{Roberta-DeductReasoner} model consistently yields higher results than baselines.
%These two datasets contain more one-step or two-step expressions (Figure \ref{fig:data_opnum}), the overall performance of our model is particularly better than the baselines.
On MathQA, our model also performs better on questions that involve 2, 3, and 4 operations.
% On MathQA, although our model's results are comparable to the baseline for questions involving 1 operation only, 
% it performs much better on questions that involve 2, 3, and 4 operations.
For the other more challenging dataset SVAMP, our model has comparable performance with the baseline on $1$-step questions, but achieves significantly better results (+$14.3$ points) on questions that involve $2$ steps.
%Though we achieve equally good performance on MathQA, we obtain better generalization on questions with 2, 3 and 4 operations.
%Our \textsc{Roberta-DeductReasoner} achieves 11 points better than Roberta-Graph2Tree under 2-step questions on SVAMP. 
Such comparisons on MathQA and SVAMP show that our model has a robust reasoning capability on more complex questions. 
%It is intuitive to say that higher performance on expressions with fewer steps leads to higher performance on expressions with more steps. 
%However, we did not see a better performance on one-step questions as the MathQA has least percentage for one-step questions. 
%Such comparison can be found in the MathQA as well. 

{\color{black}We observe that all models (including ours and existing models) are achieving much lower accuracy scores on SVAMP, as compared to other datasets.
	We further investigate the reason for this.}
\citet{patel2021are} added irrelevant information such as extra quantities in the question to confuse the models. 
We quantify the effect by counting the percentage of instances which have quantities unused in the equations. 
As we can see in Table \ref{tab:unusedquant}, SVAMP {\color{black}has the largest proportion} (i.e., $44.5$\%) of instances whose gold equations do not fully utilize all the quantities in the problem text. 
The performance also significantly drops on those questions with more than one unused quantity on all datasets. 
The analysis suggests that our model still suffer from extra irrelevant information in the question and the performance is severely affected when such irrelevant information appears more frequently. 

%\paragraph{Deductive Ability}
%not yet comet out

\begin{table}[t!]
	\centering
	\setlength{\tabcolsep}{2pt} % Default value: 6pt
	\renewcommand{\arraystretch}{1.1} % Default value: 1
	\resizebox{0.95\linewidth}{!}{
		\begin{tabular}{ccccccccc}
			\toprule
			\multirow{2}{*}{\#\textbf{Operation}} & \multicolumn{2}{c}{\textbf{MAWPS}} & \multicolumn{2}{c}{\textbf{Math23k}} & \multicolumn{2}{c}{\textbf{MathQA}}&  \multicolumn{2}{c}{\textbf{SVAMP}}\\
			& Baseline & \textsc{Ours} & Baseline & \textsc{Ours} & Baseline & \textsc{Ours} & Baseline & \textsc{Ours}  \\
			\midrule
			1 & 88.2 & \textbf{92.7} & 91.3 & \textbf{93.6} & \textbf{77.3}  & \textbf{77.4} & \textbf{51.9} & \textbf{52.0}\\
			2 & 91.3 & \textbf{91.6} & 89.3&  \textbf{92.0} & 81.3 & \textbf{83.5}& 17.8& \textbf{32.1}\\
			3 & -& - & 74.5&\textbf{77.0} & 81.9& \textbf{83.4}& -& -\\
			4 & -& - & 59.1& \textbf{60.3} & 79.3 &\textbf{81.7} & - & -\\
			>=5 & -&- &56.5 & \textbf{69.2} & \textbf{71.5} & \textbf{71.4} & -& -\\ \midrule
			\multicolumn{3}{l}{\textbf{Overall Performance}} & & & & & & \\
			Equ Acc. & 80.8 & 88.6 & 71.2 & 79.0 & 74.0 &  74.0 &40.9 &  45.0\\ 
			Val Acc. & 88.7 & 92.0 & 82.4 & 85.1 & 77.1 & 78.6 & 43.8 & 47.3\\
			%			>=6 & -&- & 57.1 & \textbf{81.1} & & 73.8& -& -\\
			\bottomrule
		\end{tabular}
	}
	\vspace{-1.5mm}
	\caption{Acc. under different number of operations.}
	\label{tab:diffops}
\end{table}

%\paragraph{Effect of Different Pre-trained Language Models} 
%Besides comparing the monolingual pre-trained models, we also evaluate the expression parsing model with multilingual pre-trained model. 
%Generally, the \textsc{ExpPar} with both monolingual and multilingual language models outperform the existing strong baselines.
%The monolingual language-based models give us a slightly better performance compared with the multilingual language-based models. 
%We did not mix up the training data of different languages to train a truly multilingual models. 
%Because mixing up different domains may need to a harmful results~\cite{tan2021investigating,li2021seeking}. 

\begin{table}[t!]
	\centering
	\resizebox{0.9\linewidth}{!}{
		\begin{tabular}{lcccc}
			\toprule
			& \textbf{MAWPS} & \textbf{Math23k} & \textbf{MathQA} & \textbf{SVAMP} \\
			Unused & 6.5\% & 8.2\% &  20.7\% & 44.5\%\\
			\midrule
			Accuracy (unused $=0$) & 93.6 & 87.1 & 81.4 & 63.6 \\
			Accuracy (unused $\geq 1$) & 100.0$\dagger$ & 62.1 & 67.4 & 27.0 \\
			\bottomrule
		\end{tabular}
	}
	\vspace{-1.5mm}
	\caption{Value accuracy with respect to the number of unused quantities. The second row shows the percentage of instances that have unused quantities. $\dagger$: may not be representative as there are only 3 instances.}
	\label{tab:unusedquant}
\end{table}

\paragraph{Effect of Rationalizer}

Table \ref{tab:rationalizer} shows the performance comparison with different rationalizers. 
As described in \Section\ref{sec:parsing}, the rationalizer is used to update the quantity representations at each step, so as to better ``prepare them'' for the subsequent reasoning process given the new context.
%such that the model is aware of previously calculated expression. 
We believe this step is crucial for achieving good performance, especially for complex MWP solving. 
% For example, the first two steps in solving the expression $(3+4)*(5+6)$ are $3+4$ and $5+6$.  
% Without quantity rationaliation, the quantity representation is never updated.
% It could be likely that we cannot obtain the $5+6$ in the second step, as the $3+4$ expression always have the largest score. 
As shown in Table \ref{tab:rationalizer}, the performance drops by $7.3$ points in value accuracy for Math23k without rationalization, confirming the importance of rationalization in solving more complex problems that involve more steps.
%MWP solving.
%\textcolor{darkgrass}{We investigate the performance and find that the accuracy on 1-step and 2-step questions are not affected, while starts to have 22 points drop for 3-step questions on Math23k.
	%This is consistent with our motivation in proposing the rationalizer. 
	%}
As most of the questions in MAWPS involve only $1$-step questions, the significance of using rationalizer is not fully revealed on this dataset.

\begin{table}[t!]
	\centering
	\resizebox{0.9\linewidth}{!}{
		\begin{tabular}{l cccc}
			\toprule
			\multirow{2}{*}{\textbf{Rationalizer}} & \multicolumn{2}{c}{\bf MAWPS} & \multicolumn{2}{c}{\bf Math23k} \\
			& \textbf{Equ Acc.} & \textbf{Val Acc.} & \textbf{Equ Acc.} & \textbf{Val Acc.}\\
			\midrule
			\textsc{None} & 88.4 & 91.8 & 71.5 & 77.8 \\
			Self-Attention & 88.3 & 91.7 & 77.5  & 84.8 \\
			GRU unit & 88.6 & 92.0 & 79.0 & 85.1 \\
			\bottomrule
		\end{tabular}
	}
	\vspace{-1mm}
	\caption{Performance comparison on different rationalizer using the Roberta-base model.}
	\label{tab:rationalizer}
\end{table}

It can be seen that using self-attention achieves  worse performance than the GRU unit.
We believe the lower performance by using multi-head attention as rationalizer may be attributed to two reasons.
First, GRU comes with sophisticated internal gating mechanisms, which may allow richer representations for the quantities.
Second, attention, often interpreted as a mechanism for measuring similarities~\cite{katharopoulos2020transformers}, may be inherently biased when being used for updating quantity representations.
This is because when measuring the similarity between quantities and a specific expression (Figure \ref{fig:rationalizer}), those quantities that have just participated in the construction of the expression may receive a higher degree of similarity.

\subsection{Case Studies}
%\subsection{Describing Intermediate Expression}
%Given the intermediate representation, we can generate the description to describe the expression in natural language. (\textcolor{red}{show BLEU score and some examples.})

\begin{figure}[t!]
	\centering
	\adjustbox{max width=0.95\linewidth}{
		\begin{tikzpicture}[node distance=2.0mm and 5.0mm, >=Stealth, 
			wordnode/.style={draw=none, minimum height=5mm, inner sep=0pt},
			chainLine/.style={line width=0.8pt,-, color=mygray},
			entbox/.style={draw=black, rounded corners, fill=red!20, dashed},
			mathop/.style={draw=deepblue, circle, minimum height=3mm, inner sep=1pt, line width=1pt},
			quant/.style={draw=deeppurple, circle, minimum height=7mm, inner sep=1pt, line width=1pt},
			expr/.style={draw=electronblue, rectangle, minimum width=20mm, minimum height=4.5mm, inner sep=4pt, line width=1pt, rounded corners},
			invis/.style={draw=none},
			]
			%			\matrix (sent) [matrix of nodes, column 2/.style={anchor=base west}, column 1/.style={anchor=base west}, nodes in empty cells, execute at empty cell=\node{\strut};]
			%			{
				%				\textbf{Question}: & \textit{\Large a car dealership has 40 cars on the lot , 15 \% of which } \\
				%				&  \textit{\Large are silver . if the dealership receives a new shipment of 80 cars , 45 \% of which are not silver , what percentage of total number of cars are silver ? } \\
				%			};
			
			\node[wordnode, align=left] (question) {\textbf{Question}: \it Xiaoli and Xiaoqiang typed a manuscript together. Their  \\\it typing speed ratio was \textbf{5}:\textbf{3}. Xiaoli typed \textbf{1,400} more  words than \\\it Xiaoqiang. How many words are there  in this  manuscript?};
			\node[wordnode, below=of question.west, anchor=west, yshift=-10mm] (gold) {\textbf{Gold Expr}: $\frac{1400}{5\div(5+3) - 3\div(5+3)}$~~~~~\textbf{Answer}: $5600$};
			
			\node[wordnode, below=of gold.west, anchor=west, yshift=-4mm] (goldstruct) {\textbf{Gold deduction}:};
			
			\node[expr] (m10) [below= of gold.west, anchor=west, yshift=-14mm] {\textcolor{black}{$5 + 3 = 8$}};
			\node[mathop, pinkgla] (st1) [below= of m10, yshift=1mm] {\scriptsize $1$}; 
			\node[expr] (m20) [right= of m10, xshift=9mm] {{\textcolor{black}{$5\div8 = 0.625$}}};
			\node[mathop, pinkgla] (st2) [below= of m20, yshift=1mm] {\scriptsize $2$}; 
			\node[expr] (m30) [right= of m20, xshift=9mm] {{\textcolor{black}{$3 \div 8 = 0.375$}}};
			\node[mathop, pinkgla] (st3) [below= of m30, yshift=1mm] {\scriptsize $3$};

			\node[expr] (m40) [below left= of m20, xshift=15mm, yshift=-5mm] {{\textcolor{black}{$0.625 - 0.375 = 0.25$}}};
			\node[mathop, pinkgla] (st4) [below= of m40, yshift=1mm] {\scriptsize $4$}; 

			\node[expr] (m50) [right= of m40, xshift=7mm, yshift=0mm] {{\textcolor{black}{$1400 \div 0.25 = 5600$}}};
			\node[mathop, pinkgla] (st5) [below= of m50, yshift=1mm] {\scriptsize $5$}; 
			
			\draw [chainLine, ->] (m10) to [] node[] {} (m20);
			\draw [chainLine, ->] (m10) to [out=20, in=160, looseness=0.8] node[] {} (m30);
			\draw [chainLine, ->] (m30) to [] node[] {} (m40);
			\draw [chainLine, ->] (m20) to [] node[] {} (m40);
			\draw [chainLine, ->] (m40) to [] node[] {} (m50);

			\node[wordnode, below=of m10.west, anchor=west, yshift=-20mm] (predstruct) {\textbf{Predicted deduction}:};
			\node[expr] (p10) [below= of m10.west, anchor=west, yshift=-28mm] {\textcolor{black}{$5 - 3 = 2$}};
			\node[mathop, pinkgla] (st1s) [below= of p10, yshift=1mm] {\scriptsize $1$}; 
			\node[expr] (p20) [right= of p10, xshift=0mm] {{\textcolor{black}{$1400 \div 2 = 700$}}};
			\node[mathop, pinkgla] (st2s) [below= of p20, yshift=1mm] {\scriptsize $2$}; 
			\node[expr] (p30) [right= of p20, xshift=-2mm] {{\textcolor{black}{$ 5 + 3 = 8$}}};
			\node[mathop, pinkgla] (st3s) [below= of p30, yshift=1mm] {\scriptsize $3$}; 
			
			\node[expr] (p40) [right= of p30, xshift=0mm] {{\textcolor{black}{$700 \times 8 = 5600$}}};
			\node[mathop, pinkgla] (st4s) [below= of p40, yshift=1mm] {\scriptsize $4$}; 
			
			\draw [chainLine, ->] (p10) to [] node[] {} (p20);
			\draw [chainLine, ->] (p20) to [out=20, in=160, looseness=0.8] node[] {} (p40);
			\draw [chainLine, ->] (p30) to [] node[] {} (p40);

		\end{tikzpicture}
	}
	\vspace{-2mm}
	\caption{Deductive steps by our reasoner.}
	\label{fig:example_expression}
\end{figure}

%\subsection{Quantity Coverage}

\paragraph{Explainability of Output}

%We elaborate from the results to show that the deductive reasoner provides explainable prediction.
Figure \ref{fig:example_expression} presents an example prediction from Math23k.
In this question, the gold deductive process first obtains the speed difference by ``$5\div (5+3) - 3\div (5+3)$'' and the final answer is $1400$ divided by this difference.
On the other hand, the predicted deductive process offers a slightly different understanding in speed difference.
Assuming speed can be measured by some abstract ``units'', the predicted deductive process first performs subtraction between $5$ and $3$, which gives us ``$2$ units'' of speed difference.
%(``$2$ units of speed'').
Next, we can obtain the number of words associated with each speed unit ($1400\div 2$).
Finally, we can arrive at the total number of words by multiplying the number of words per unit ($700$) and the total number of units ($8$).\footnote{Interestingly, when we presented this question to 3 human solvers, 2 of them used the first approach and 1 of them arrived at the second approach.}
Through such an example we can see that our deductive reasoner is able to produce explainable steps  to understand the answers.

\begin{figure}[t!]
	\centering
	\adjustbox{max width=0.95\linewidth}{
		\begin{tikzpicture}[node distance=2mm and 5.0mm, >=Stealth, 
			wordnode/.style={draw=none, minimum height=5mm, inner sep=0pt},
			chainLine/.style={line width=1.1pt,-, color=deepblue},
			entbox/.style={draw=black, rounded corners, fill=red!20, dashed},
			mathop/.style={draw=deepblue, circle, minimum height=3mm, inner sep=1pt, line width=1pt},
			quant/.style={draw=deeppurple, circle, minimum height=7mm, inner sep=1pt, line width=1pt},
			expr/.style={draw=deepblue, rectangle, minimum width=20mm, inner sep=2pt, line width=1pt, rounded corners},
			invis/.style={draw=none},
			]
			%			\matrix (sent) [matrix of nodes, column 2/.style={anchor=base west}, column 1/.style={anchor=base west}, row sep=0.001mm, nodes in empty cells, execute at empty cell=\node{\strut};]
			%			{
				%					\textbf{Question}: & \textit{There are \textbf{255} apple trees in the orchard. Planting another  } \\
				%					&  \textit{\textbf{35} pear trees makes them equal.  If every \textbf{20} pear trees} \\
				%				};
			
			\node[wordnode, align=left] (question) {\textbf{Question}: \it There are \textbf{255} apple trees in the orchard.  \color{red}{Planting another}\\\it \textcolor{red}{\textbf{35} pear trees makes the number exactly the same as the apple trees.} If \\\it every \textbf{20} pear trees   are planted in a row,  how many  rows can be  planted\\\it in total?};
			\node[wordnode, below=of question.west, anchor=west, yshift=-10mm] (gold) {\textbf{Gold Expr}: $(255-35)\div20$~~~~~\textbf{Answer}: $11$};
			\node[wordnode, below=of gold.west, anchor=west, yshift=-3mm] (pred) {\textbf{Predicted Expr}: $(\textcolor{red}{255 + 35})\div20$~~~~~\textbf{Predicted}: $14.5$};

			\node[wordnode, below=of gold.west, anchor=west, yshift=-8mm] (predstruct) {\textbf{Deductive Scores:}};
			\node[expr] (p10) [below= of predstruct.west, anchor=west, yshift=-5mm, label=right:{Prob.: $0.068 >$}] {\textcolor{black}{$255 + 35= 290$}};
			
			\node[expr,right= of p10, xshift=22mm, label=right:{Prob.: $0.062$}] (p20) [] {\textcolor{black}{$255 - 35= 220$}};
			
			\node[wordnode, below=of p10.west, anchor=west, yshift=-11mm, align=left] (question1) {\textbf{Perturbed Question}: \it There are \textbf{255} apple trees in the orchard.  \color{blue}{\textbf{The} }\\\it \textcolor{blue}{\textbf{number of pear trees is 35 fewer than the apple trees.}} If every \textbf{20} pear \\\it trees   are planted in a row,  how many  rows can be  planted in total?};
			
			\node[expr] (pp10) [below= of question1.west, anchor=west, yshift=-8mm, label=right:{ \textbf{\textcolor{blue}{Prob.: $0.061 <$}}}] {\textcolor{black}{$255 + 35= 290$}};
			
			\node[expr,right= of pp10, xshift=22mm, label=right:{ \textbf{\textcolor{blue}{Prob.: $0.067$}}}] (p20) [] {\textcolor{black}{$255 - 35= 220$}};
			
		\end{tikzpicture}
	}
	\vspace{-2mm}
	\caption{Question perturbation in deductive reasoning.}
	\label{fig:perturbation}
\end{figure}

\paragraph{Question Perturbation}
The model predictions also give us guidance to understand the errors.
%help us better
Figure \ref{fig:perturbation} shows how we can perturb a question given the error prediction (taken from Math23k).
As we can see, the first step is incorrectly predicted with the ``$+$'' relation between $255$ and $35$.
Because the first step involves the two quantities in the first two sentences, where we can locate the possible cause for the error.
The gold step has a probability of $0.062$ which is somewhat lower than the incorrect prediction.
We believe that the second sentence (marked in red) may convey  semantics that can be challenging for the model to digest, resulting in the incorrect prediction.
Thus, we perturb the second sentence to make it semantically more straightforward {\color{black}(marked below in blue)}.
% We can see that after the purtubation, 
The probability for the sub-expression $225-35$ becomes higher after the purtubation, leading to a correct prediction (the ``$-$'' relation).
Such an analysis demonstrates the strong interpretability of our deductive reasoner, and highlights the important connection between math word problem solving and reading comprehension, a topic that has been studied in educational psychology \cite{vilenius2008association}.
%Figure \ref{fig:perturbation} two some incorrect predicted samples and the corresponding expressions. 
%As we can see from the first example, the expression is wrongly predicted at the second step, where . 
%We perform perturbation on the sentence that contain the corresponding quantity $q_3$, to make the description more accurate. 
%After the perturbation, w

\subsection{Practical Issues}
\label{sec:discussion}
% \paragraph{Scalability}

We discuss some practical issues with the current model in this section.
Similar to most previous research efforts~\cite{li2019modeling,xie2019goal}, our work needs to maintain a list of constants (e.g., $1$ and $\pi$) as additional candidate quantities. 
However, a large number of quantities could lead to a large search space of expressions (i.e., $\mathcal{H}$).
In practice, we could select some top-scoring quantities and build expressions on top of them~\cite{lee2018higher}.
Another assumption of our model, as shown in Figure \ref{fig:model}, is that only binary operators are considered.
Actually, extending it to support unary or ternary operators can be straightforward.
Handling unary operators would require the introduction of some unary rules, and a ternary operator can be defined as a composition of two binary operators.

Our current model performs the greedy search in the training and inference process, which could be improved with a beam search process.
One challenge with designing the beam search algorithm is that the search space $\mathcal{H}^{(t)}$ is expanding at each step $t$ (Equation \ref{equ:loss}).
We empirically found the model tends to favor outputs that involve fewer reasoning steps.
In fact, better understanding the behavior and effect of beam search in seq2seq models remains an active research topic \cite{cohen2019empirical,koehn2017six,hokamp2017lexically}, and we believe how to perform effective beam search in our setup could be an interesting research question that is worth exploring further.

% We leave the issues above for future work to design a more scalable model and  beam search algorithm for inference.

\section{Conclusion and Future Work}

We provide a new perspective to the task of MWP solving and argue that it can be fundamentally regarded as a {complex relation extraction} problem.
Based on this observation, and motivated by the deductive reasoning process, we propose an end-to-end deductive reasoner to obtain the answer expression in a step-by-step manner.
At each step, our model performs iterative mathematical relation extraction between quantities.
Thorough experiments on four standard datasets demonstrate that our deductive reasoner is robust and able to yield new state-of-the-art performance.
The model achieves particularly better performance for complex questions that involve a larger number of operations.
It offers us the flexibility in interpreting the results, thanks to the deductive nature of our model.

% One advantage of our approach is its convenience in incorporating declarative knowledge in reasoning process. 
Future directions that we would like to explore include how to effectively incorporate commonsense knowledge into the deductive reasoning process, and how to facilitate counterfactual reasoning~\cite{richards1999role}.
% We are also interested in building structured models to 
% {\color{red}TODO: to add future work. some problems require background knowledge, children needs visual/imagination for solving maths problems.}
%The underlying model architecture allows us to interpret the deductive process for obtaining the intermediate expressions and the final answer.
%We further analyze the deductive procedure during inference is explainable and provide us with guidance to understand the predictions.

%In the future, we plan to ...

%As the overall performance of \textsc{DeductReasoner} on SVAMP~\cite{patel2021are} is still low, we plan to perform more fine-grained reasoning beyond only the quantities. 


\section*{Acknowledgements}

We would like to thank the anonymous reviewers and our ARR action editor for their constructive comments, and Hang Li for helpful discussions and comments on this work.
%This project/research is partially supported by the National Research Foundation, Singapore under its AI Singapore Programme (AISG Award No: AISG-RP-2019-012).
This work was done when Jierui Li was working as a research assistant at SUTD, and when Wei Lu was serving as a consultant at ByteDance AI Lab.

\bibliography{parsing_mwp}
\bibliographystyle{acl_natbib}

\appendix

\section{Importance of Rationalizer}
\label{sec:appendix_rationalizer}
We further elaborate on the importance of the rationalizer module in this section.
As mentioned in \Section \ref{sec:parsing}, it is crucial for us to properly update the quantity representations, especially for questions that require more than 3 operations to solve (i.e., $t\geq 3$).
Because if the quantity representations do not get updated as we continue the deductive reasoning process, those expressions that were initially highly ranked (say, at the first step) would always be preferred over those lowly ranked ones throughout the process. 

We provide an example here to illustrate the scenario.
Suppose our target expression is $(1+2)*(3+4)$, the first step is to predict:
\begin{equation}
	e^{(1)} = 1 + 2
\end{equation}

In order to obtain the correct intermediate expression $1+2$ as $e^{(1)}$, the model has to give the highest score to this expression.
Note that, the score of the expression $e_{1,2,+}$ also has to be larger than the score of $e_{3,4,+}$.
\begin{equation}
	s(e_{1,2,+}^{(1)}) > s(e_{3,4,+}^{(1)})
\end{equation}

However, in order to reach the final target expression, in the next step, the model needs to construct the intermediate expression $3+4$.
Without the rationalizer, the representations for the quantities are unchanged, so we would have:
\begin{equation}
	s(e_{1,2,+}^{(2)}) = s(e_{1,2,+}^{(1)}) > s(e_{3,4,+}^{(1)}) = s(e_{3,4,+}^{(2)})
	\label{eqn:rr}
\end{equation}

From here we could see that the model would not be able to produce the intermediate expression $3+4$ in the second step (but would still prefer to generate another $1+2$).
With the rationalizer in place, the above two equations in Equation \ref{eqn:rr} in general may not hold, which effectively prevents such an issue from happening.

\section{Additional Implementation Details}
We implement our model with PyTorch and run all experiments using Tesla V100 GPU.
The feed-forward network in our model is simply linear transformation followed by the ReLU activation. 
We also apply layer normalization and dropout in the feed-forward network. 
The hidden size in the feed-forward network is 768, which the is same as the hidden size used in BERT/Roberta.

\section{Detailed Comparison on Math23k Dataset}
\label{sec:math23k_detailed}
We try to find the experiment details of previous work on the Math23k dataset. 
\begin{table*}[t!]
	\centering
	\resizebox{1.0\linewidth}{!}{
		\begin{tabular}{lcccccl}
			\toprule
			\multirow{3}{*}{\bf Model}& & \multicolumn{4}{c}{\bf ``Split Variant''}& \multirow{3}{*}{\bf Remark} \\
			& \bf Beam& \bf Train/Val/Test & \bf Train/Test & \bf 5-fold  & \bf Customized &  \\
			& \bf Size & \bf 21162/1000/1000 & \bf  22162/1000 & \bf CV& \bf Split & \\
			\midrule
			Group Attention~\cite{li2019modeling} & 5 & - & 69.5 & 66.9 &- & \\
			GTS~\cite{xie2019goal} & 5 & - & - & 74.3 &- & \\
			KA-S2T~\cite{wu2020knowledge} & 5 & - & - & - & 76.3 & They randomly split into 80\%/20\% \\
			MultiE\&D~\cite{shen2020solving} & 5 & \multicolumn{2}{c}{78.4 \footnotesize\it (unknown)$\dagger$} & 76.9 & & \\
			Graph2Tree~\cite{zhang2020graph} & 5 & 77.4 & & 75.5 &- & \\
			NeuralSymbolic~\cite{qin2021neural} & 1 &- &- & 75.7 & -& \\
			NUM2ST~\cite{wu2021math} & 5 &- & -&- & 78.1 & They randomly split into train/val/test \\
			HMS~\cite{lin2021hms} & 1 & 76.1 &- &- &- & \\
			BERT-Tree~\cite{li2021seeking} & 3 & 82.4 &- & -&- & \\
			mBART-Large~\cite{shen2021generate} & 10 & \multicolumn{2}{c}{85.4 \footnotesize\it (unknown)$\dagger$}$\dagger$ & 84.3 &- & \\
			\midrule
			\textsc{Roberta-DeductReasoner} & 1 & 84.3 {\small{($\pm$ 0.34)}} & 86.0 {\small{($\pm$ 0.26)}} & 83 {\small{($\pm$ 0.36)}} &- & \\
			\textsc{Roberta-Large-DeductReasoner}  & 1 & 85.8 {\small{($\pm$ 0.42)}} & 87.1 {\small{($\pm$ 0.21)}} & - &  - & \\
			\bottomrule
		\end{tabular}
	}
	\vspace{-1mm}
	\caption{Detailed comparison of different approaches on the Math23k dataset. $\dagger$ indicates the paper only mentioned the results are evaluated on test set but not mentioned if they used validation set for experiment.}
	\label{tab:math23k_additional_exp}
\end{table*}
Table \ref{tab:math23k_additional_exp} shows their performance with respect to different data splits, different beam sizes, etc.

\squishlist
\item Group Attention~\cite{li2019modeling}: According to their paper in Table 2, they use the train/test split. 
\item GTS~\cite{xie2019goal}: They only report the five-fold cross-validation performance. 
\item KA-S2T~\cite{wu2020knowledge}: According to \Section 3.1 in their paper, they use their customized split though they still directly compared with the GTS approach. 
\item MultiE\&D~\cite{shen2020solving}: They did not mention that if they have validation set. But they mentioned that the results are evaluated on the test set. Thus, we marked it ``\textit{unknown}''. 
\item Graph2Tree~\cite{zhang2020graph}: They also mentioned that the results are evaluated on test set. But the validation set could be used in their implementation based on the observation in their codebase.
\item NeuralSymbolic~\cite{qin2021neural}: They used greedy search (\Section 4.2.2) for generation and experimented with 5-fold cross validation in the experiments (\Section 4.3).
\item NUM2ST~\cite{wu2021math}: They used customized splits according to \Section 3.1 in their paper. 
\item HMS~\cite{lin2021hms}: The beam size is not mentioned in the paper. But we found the default value in the open-source codebase is 1. They did not mentioned the exact split either but simply mentioning ``\textit{follow previous work}''. 
We assume they are using the train/validation/test split as we found that the codebase contains the validation set.
\item BERT-Tree~\cite{li2021seeking}: They use train/validation/test split as described in the paper.
\item mBART-Large~\cite{shen2021generate}: They did not mentioned the split either but mentioning that the results are evaluated on the test set. 
\squishend

We present our performance for all the settings in Table \ref{tab:math23k_additional_exp}.
To compare with the recent work using mBart-Large~\cite{shen2021generate}, we also report the performance with a Roberta-Large encoder for our \textsc{DeductReasoner} on the test set.

\end{document}